\pdfoutput=1

\documentclass[11pt]{article}

\usepackage{ACL2023}

\usepackage{times}
\usepackage{latexsym}

\usepackage[T1]{fontenc}

\usepackage[utf8]{inputenc}

\usepackage{microtype}

\usepackage{inconsolata}

\usepackage{algorithm}
\usepackage{algorithmic}
\usepackage{adjustbox}
\usepackage{bbold}
\usepackage{kotex}
\usepackage{graphicx,subcaption}
\usepackage{colortbl}
\usepackage{tcolorbox}
\usepackage{placeins}

\usepackage{multirow}
\usepackage{booktabs}
\usepackage{enumitem}
\title{Investigating the Influence of Prompt-Specific Shortcuts in AI Generated Text Detection}

\author{
    Choonghyun Park\textsuperscript{\rm 1},
    Hyuhng Joon Kim\textsuperscript{\rm 1},
    Junyeob Kim\textsuperscript{\rm 1},
    Youna Kim\textsuperscript{\rm 1},\\
    \textbf{Taeuk Kim\textsuperscript{\rm 2},
    Hyunsoo Cho\textsuperscript{\rm 3},
    Hwiyeol Jo\textsuperscript{\rm 4},
    Sang-goo Lee\textsuperscript{\rm 1},
    Kang Min Yoo\textsuperscript{\rm 1 5 6}}\\
    \textsuperscript{\rm 1}Seoul National University,
    \textsuperscript{\rm 2}Hanyang University,
    \textsuperscript{\rm 3}Ewha Womans University,\\
    \textsuperscript{\rm 4}NAVER Search US,
    \textsuperscript{\rm 5}NAVER AI LAB,
    \textsuperscript{\rm 6}NAVER Cloud\\
    \{pch330, heyjoonkim, juny116, anna9812, sglee\}@europa.snu.ac.kr\\
    kimtaeuk@hanyang.ac.kr, chohyunsoo@ewha.ac.kr\\
    \{hwiyeol.jo, kangmin.yoo\}@navercorp.com
}

\begin{document}
\maketitle

\def\attackMethod{FAILOpt}
\def\attackMethodLong{Feedback-based Adversarial Instruction List Optimization}
\def\attackMethodFull{\attackMethodLong\space(\attackMethod)}

\begin{abstract}

\label{sec:abstract}
AI Generated Text (AIGT) detectors are developed with texts from humans and LLMs of common tasks. Despite the diversity of plausible prompt choices, these datasets are generally constructed with a limited number of prompts. The lack of prompt variation can introduce prompt-specific shortcut features that exist in data collected with the chosen prompt, but do not generalize to others. In this paper, we analyze the impact of such shortcuts in AIGT detection.
We propose \attackMethodFull, an attack that searches for instructions deceptive to AIGT detectors exploiting prompt-specific shortcuts. \attackMethod\ effectively drops the detection performance of the target detector, comparable to other attacks based on adversarial in-context examples. We also utilize our method to enhance the robustness of the detector by mitigating the shortcuts. Based on the findings, we further train the classifier with the dataset augmented by \attackMethod\ prompt. The augmented classifier exhibits improvements across generation models, tasks, and attacks. Our code will be available at \url{https://github.com/zxcvvxcz/FAILOpt}.
    
\end{abstract}

\section{Introduction}
\label{sec:introduction}

Large Language Models (LLMs)~\cite{achiam2023gpt,anthropic2024claude,touvron2023llama} marked a phenomenal advancement in natural language processing (NLP). The capacity of these models to write human-level texts, and adapt to new tasks through prompting, makes them exceptionally beneficial tools for various fields. Meanwhile, there is also a rising concern about misuse. Students can submit generated answers as if their own~\cite{bohacek2023unseen,busch2023too}, and malignant users can use them to spread misinformation~\cite{pan2023risk,spitale2023ai}.

\begin{figure}[t!]
\includegraphics[width=1.02\linewidth]
{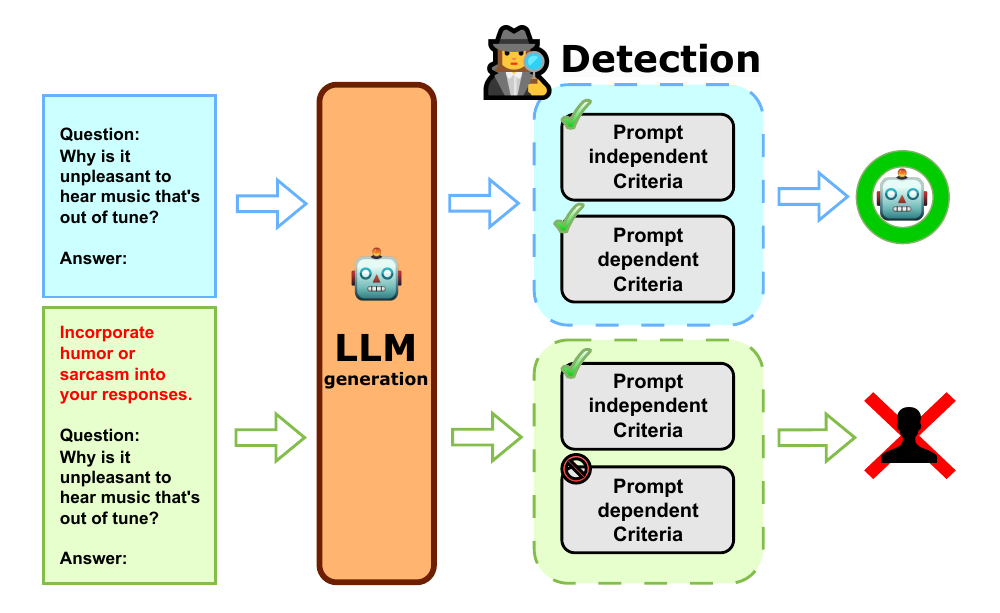}
\centering
\caption{An illustration of the detection failure caused by the reliance on prompt-specific shortcuts.}
\label{fig:shortcut_situation}
\end{figure}

This threat put a spotlight to the development of AI Generated Text (AIGT) detectors that can tell if a given text is written by AI or human. Recent works proposed detection approaches for the authorship, and achieved promising results on experimental settings~\cite{guo-etal-2023-hc3,mitchell2023detectgpt,su2023detectllm,koike2024outfox,tulchinskii2023intrinsic}. However, following works revealed that these detectors can be deceived effectively via adversarial attacks. These works provide practical attack scenarios harmful to detection performances but do not provide insights into the sources of such vulnerabilities. 

This paper investigates one plausible reason behind this failure: shortcut learning of prompt-specific shortcuts. 
Shortcut learning \cite{geirhos2020shortcut,hermann2024on} refers to the phenomenon where a model learns to rely on shortcuts, the spurious cues that show correlations of inputs and labels in train data but cannot be applied to real-world scenarios. A common example is image classifiers that leverage background in the inputs to discriminate different objects. Various works on NLP \cite{du2023shortcut} show that language models are also subject to such issues. Shortcut learning makes detectors unreliable for practical uses, and it is important to train models on balanced data that correctly represent the data domain. 

Previous literature trained and evaluated AIGT detectors on datasets constructed with human and AI generated texts for common inputs~\cite{guo-etal-2023-hc3,su2023hc3,chen2023gpt}. For each task, it is common to collect human text with corresponding AI texts that share the same input. 
Despite the variety of applicable prompts, only a small number of prompts are considered in these works.
As recent LLMs show high instruction-following capacity, the limited prompt diversity can introduce shortcuts specific to the generations from the data collection prompt. Figure \ref{fig:shortcut_situation} illustrates the danger of prompt-specific shortcuts in AIGT detection. Attack results based on adversarial in-context examples~\cite{lu2024large,shi2023red} and recent analysis on the influence of prompts~\cite{koike2023you,zhang2023assaying} in detection performance also show the importance of prompts in AIGT detection. 
 
In this paper, we term such shortcuts as prompt-specific shortcuts and show their harmful influence on the development of AIGT detectors. 
To this end, we first show that the performance of an AIGT detector trained with generations from limited prompts depends on prompt-specific features, while other detectors do not rely on them. We propose an attack method named \attackMethodFull\ to find a list of instructions that ask the LLM to alter prompt-specific features of its generations that a detector relies on. Experiments on multiple datasets show that generations based on the \attackMethod\ instructions are effective at eluding the detector, but such influence diminishes on other detectors not trained on the same data.
Second, we find that the mitigation of such shortcuts enhances the general robustness of a detector. As we additionally train a vulnerable detector on the augmented data composed of AIGTs from base prompt and a \attackMethod\ prompt, the detection score generally increases across generation models, tasks, and attack methods.

In summary, our contributions are as follows:
\begin{itemize}[noitemsep,topsep=1pt]
    \item We confirm that developing AIGT detectors with AIGTs from limited prompts, a common setting for AIGT detection, can severely harm the robustness of detectors as they learn prompt-specific shortcuts. We support the idea with two observations: 1) We can find instructions that deteriorate the performance of a detector by perturbing prompt-specific features. 2) Training a vulnerable detector with generations based on deceptive instructions relevant to shortcuts can improve its robustness.
    \item We propose \attackMethodFull, a novel attack method that finds deceptive instructions that perturb features related to prompt-specific shortcuts. The attack \attackMethod\ achieved comparable performance.
    \item We find that \attackMethod\ can also be utilized to improve the robustness of a detector. Additional training with AIGTs from a \attackMethod\ prompt drastically improved its performance against \attackMethod. Moreover, this improvement generalizes to different generation models, tasks, and attack methods. 
\end{itemize}

\section{Related Works}
\label{sec:related works}

\subsection{Sensitivity of LLMs towards Prompt Choices}
\citet{brown2020language} reveal that LLMs are easily applicable to a new task with natural language task descriptions called prompts. Following this groundbreaking discovery, numerous prompting methodologies are studied~\cite{sahoo2024systematic,li2023large}. Recent works discovered that LLMs are sensitive to prompt designs. Variations in the design, i. e. paraphrases~\cite{zhou2022large,fernando2023promptbreeder}, order of in-context examples~\cite{lu-etal-2022-fantastically}, or formats~\cite{sclar2023quantifying} can heavily impact the accuracy of LLMs.

This leaves a significant threat in AIGT detection. A reliable detector should detect AI generations regardless of the generation prompt. Among the plausible options, there might be smart prompts that deceive detectors. Several papers mention this  issue~\cite{mitchell2023detectgpt,kirchenbauer2023watermark}, but do not analyze it deeper. Also, existing datasets to train and evaluate AIGT detectors~\cite{guo-etal-2023-hc3,chen2023gpt} are commonly constructed with generations from a single manual prompt. 

Recently, \citet{koike2023you} raises a concern on this topic, showing that AIGT detectors become unstable when LLMs are given manually written task-oriented constraints.
\citet{taguchi2024impact} analyzes multiple metric-based detectors, finding that providing the generation prompts significantly affects their performances. 
In this paper, we take a further step and point out the causal relationship between the biases from the data construction prompts and the vulnerabilities of an AIGT detector. 

\subsection{AIGT Detectors \& Attacks}
There are three prevailing types of AIGT detectors: watermark detectors, metric-based detectors, and supervised classifiers.
Watermark detectors~\cite{kirchenbauer2023watermark,kuditipudi2023robust} identify watermarks inserted in the generation phase. We do not test them in this paper as their relevance to prompt-specific features is unclear.
Metric-based detectors~\cite{mitchell2023detectgpt,su2023detectllm,bao2023fast,hans2024spotting} leverage statistical criteria that explain the difference between AI and human to detect AIGTs in a zero-shot manner.
Supervised classifiers~\cite{guo-etal-2023-hc3,chen2023gpt,huang2024ai} are trained with labeled datasets of AIGTs and human writings. 

Various effective attacks, i.e. paraphrasing the output directly~\cite{krishna2023paraphrasing,sadasivan2023aigenerated}, paraphrasing the input~\cite{ha2023black,shi2023red}, and concatenating deceptive in-context examples to the input~\cite{shi2023red,lu2024large} could drop the detection scores. 
These works show the existence of vulnerabilities but do not reveal their sources. Meanwhile, our goal is to verify the reason behind the weaknesses related to the data collection process. To this end, we design an attack 
 that suits better in analyzing prompt-specific features, and utilize it to provide enhance robustness of AIGT detectors.

\section{Overview}
\label{sec:overview_detection}
\subsection{LLM-based Generation}
LLM is an autoregressive language model that generates a text based on input texts. 
In this paper, we focus on a practical setup where an LLM generation $g_{LLM}$ is formulated as $g_{LLM}=G(t, a, x)$. $G$ represents the LLM generation function that outputs a text $g_{LLM}$ from the input text, where $t$ describes the main task, $a$ refers to an additional prompt for output alignment, and $x$ refers to the main instance that specifies the content of current input. For example, when the input is \texttt{"Question: Why is it unpleasant to hear music that's out of tune? Answer:"}, \texttt{"Question: ... Answer:"} is $t$, \texttt{"Why is it unpleasant to hear music that's out of tune?"} is $x$, and $a$ is not included in the example. Available options for $a$ include adjusting the tone (\texttt{"Answer friendly."}), assigning a persona (\texttt{"You are a helpful chatbot."}), etc. 

\subsection{AIGT Detection}
Given a text sequence $g$ written by either human or an LLM, AIGT detectors predict its score $f(g)$, which represents the likelihood of $g$ to be an AIGT. Based on the score, we assign a classification label as $y=\mathbb{1}(f(g) \geq \tau)$, where $\tau$ is the predetermined detection threshold. Note that the inputs for LLM, i.e. $t, a, x$, are not available to detectors. A reliable detector should be able to find the correct label independent of the input choices.

\subsection{Shortcut Learning}
AIGT detectors are developed with a dataset consisting of human and AI responses from the same input. It is common to use AIGT datasets constructed with AIGTs from a single $t, a$ for each task, only focusing on the variation of $x$. However, $t$ and $a$ are also important factors for generation. Even when the task is fixed, there are practically an infinite number of possible input variations. Therefore, such selection bias is likely to cause spurious correlations, deteriorating robustness of detectors as they depend on non-robust shortcuts that only explain the behavior of LLM on a subset of possible prompts. We investigate this issue with attack and defense utilizing prompt-specific shortcut features in train data.

\begin{figure*}[t!]
\centering
\includegraphics[width=0.9\textwidth]{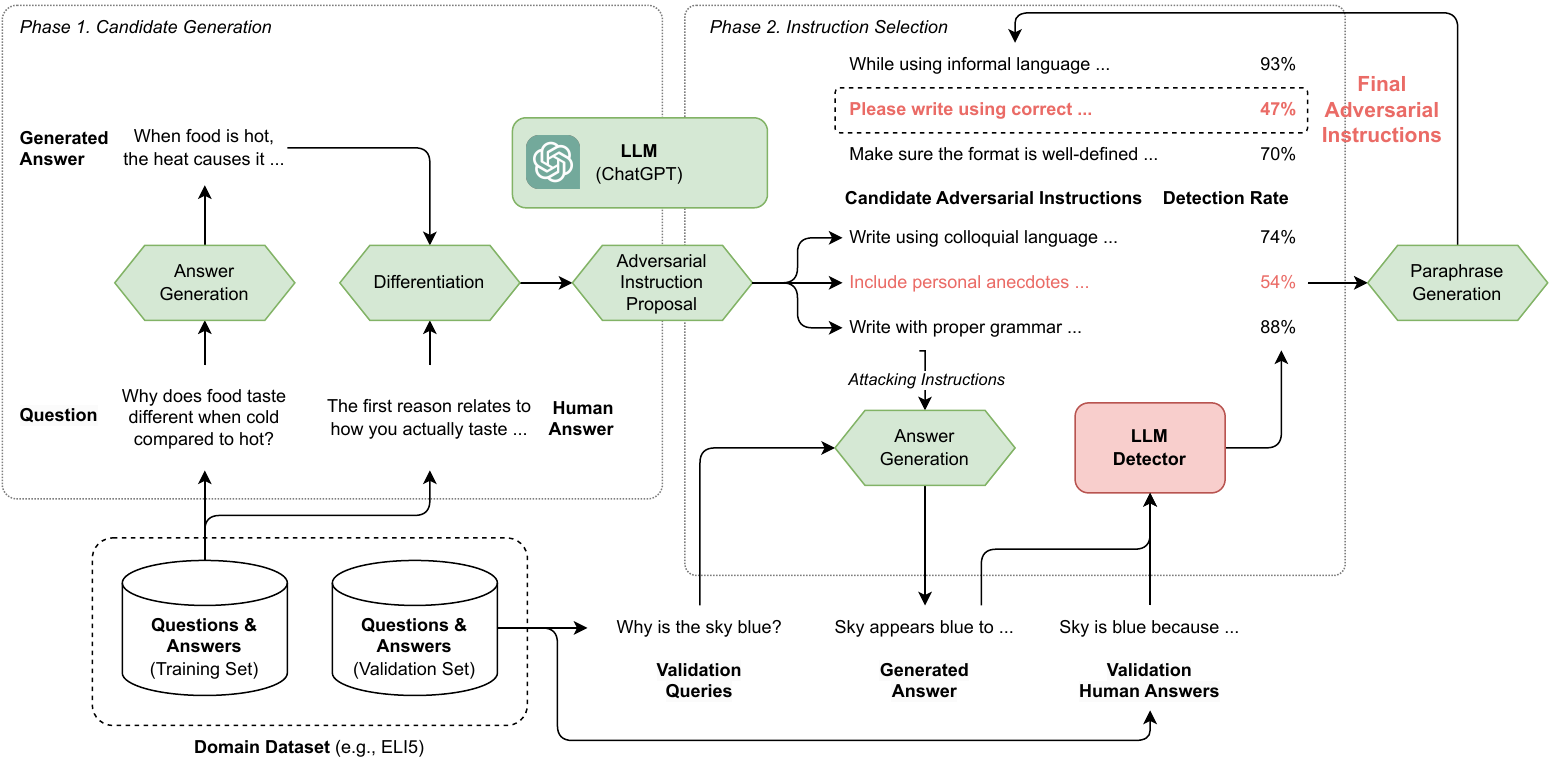}
\caption{An illustration of the first iteration of \attackMethodLong\ on ELI5. 
}
\label{fig:apa_framework}
\end{figure*}
\section{Eluding Detectors via Prompt-Specific Shortcut Exploitation}
\label{sec:instruction_based_attack}
To verify the significance of prompt-specific shortcuts in AIGT detection, we need a tool to show the vulnerability of AIGT detectors by exploiting such shortcuts. 
Recent attack works~\cite{krishna2023paraphrasing,lu2024large,shi2023red} revealed vulnerabilities in AIGT detectors, but they are not specialized in exploiting the vulnerabilities of our interest. Therefore, we propose \attackMethodFull, an attack that explicitly targets the prompt-specific shortcuts relevant to the generation prompt to deceive detectors.
\subsection{Design Outline}
\label{sec:design_outline}
\attackMethod\ leverages prompt-specific shortcuts the detectors learned to find deceptive instructions. To ensure the connection between the resulting instructions from \attackMethod\ and the shortcuts of detectors, we design \attackMethod\ to find instructions that meet two requirements. 
First, each instruction should affect representative features of the AIGTs compared to human writings. 
Second, the generations based on the additional instructions should be able to deceive detectors. 
We design the optimization process of \attackMethod\ to meet both of them.

\begin{table*}[t]
\centering
\small
\setlength{\tabcolsep}{6pt} 
\renewcommand{\arraystretch}{1} 
\begin{tabular}{cl|ccc|ccc|ccc} 
\toprule
\multicolumn{2}{c|}{\multirow{2}{*}{}} & \multicolumn{3}{c|}{ChatGPT detector} & \multicolumn{3}{c|}{Perplexity} & \multicolumn{3}{c}{DetectGPT} \\
\multicolumn{2}{c|}{} & ELI5 & XSum & SQuAD & ELI5 & XSum & SQuAD & ELI5 & XSum & SQuAD \\ 
\midrule
\multirow{6}{*}{\begin{tabular}[c]{@{}c@{}}AUROC\\($\downarrow$)\end{tabular}} & N/A & 93.33 & 80.70 & 96.08 & 97.88 & 93.00 & 98.77 & 91.39 & 80.18 & 94.34 \\
 & PARA & 89.31 & 63.61 & 83.13 & 89.94 & \textbf{78.00} & \textbf{94.15} & \textbf{78.29} & 61.99 & \textbf{85.57} \\
 & DIPPER & 92.27 & 80.71 & 84.84 & 93.66 & 81.85 & 94.60 & 81.99 & 67.52 & 87.79 \\
 & SICO & \textbf{77.21} & 50.31 & \textbf{61.45} & 97.52 & 90.44 & 94.98 & 90.55 & 75.22 & 88.90 \\
 & IP & 89.63 & \textbf{40.47} & 75.51 & 95.51 & 72.53 & 97.43 & 87.39 & \textbf{58.71} & 92.22 \\
 & \attackMethod & 78.17 & 64.92 & 88.31 & \textbf{89.36} & 87.72 & 97.70 & 87.09 & 75.34 & 91.89 \\ 
\midrule
\multirow{5}{*}{\begin{tabular}[c]{@{}c@{}}ASR\\($\uparrow$)\end{tabular}} & PARA & 14.67 & 34.59 & 27.80 & \textbf{34.35} & 42.86 & \textbf{25.30} & \textbf{33.90} & 32.24 & 27.19 \\
 & DIPPER & 20.08 & 20.52 & 40.10 & 12.86 & 48.31 & 21.97 & 30.77 & 34.63 & \textbf{30.23} \\
 & SICO & 33.92 & 53.27 & \textbf{53.94} & 5.11 & 19.33 & 22.62 & 16.24 & 17.06 & 27.05 \\
 & IP & 18.81 & \textbf{64.87} & 32.80 & 7.95 & \textbf{54.34} & 10.88 & 16.47 & \textbf{45.48} & 16.65 \\
 & \attackMethod & \textbf{46.55} & 42.31 & 19.18 & 26.73 & 27.36 & 9.18 & 19.40 & 19.66 & 19.34 \\
\bottomrule
\end{tabular}
\caption{Detection performances on attacked ChatGPT (gpt-3.5-turbo-0301) generations. We present each score in percentage. The N/A on AUROC represents the average AUROC of non-attack generations measured in the 5 attack methods. The best attack score for each column is represented in bold.}
\label{tab:attack_0301}
\end{table*}
\subsection{\attackMethodFull}
\label{sec:apa_implementation}

\attackMethod\ is an automatic attack algorithm that iteratively optimizes a list of deceptive sub-task instructions against a target detector. In each iteration, it utilizes the instruction-following capacity of LLMs to add an instruction that reduces the distinctive features of LLM generations in the list. 

Each step of \attackMethod\ consists of two phases. In the first phase, candidate generation, the model analyzes the differences between the current outputs of LLM and human writings for common input instances and generates candidate sub-task instructions that guide the LLM to generate human-like texts without changing the main task. As all candidates are relevant to the characteristics of AIGTs from the current prompt, this phase ensures to fulfill the first requirement in \ref{sec:design_outline}. 

In the second phase, instruction selection, the model evaluates the deceptive effect of each candidate, finding top-k instructions that elude a target AIGT detector. This phase ensures to fulfill the second requirement. We provide the pseudo code of \attackMethod\ in Algorithm \ref{alg:attack_algorithm} and \ref{alg:evaluation_and_select}, and the prompts for each step in Table \ref{tab:prompt_format_attack}.

\paragraph{Candidate Generation}
Given a collection of pairs of input instance $x$ and human answer $h$, $D_{tr}={(x_{tr}^{1}, h_{tr}^{1}), ..., (x_{tr}^{|D_{tr}|}, h_{tr}^{|D_{tr}|})}$, we randomly samples a batch of pairs $B_{tr}$. We ask an LLM to generate a response for each input in $B_{tr}$, and conduct several tasks to find adversarial instruction candidates based on the responses. First, the LLM compares human writings to these responses, and provides feedback as a list of $N_{feed}$ general differences between them. Each item in the list is converted into an instruction that orders the model to adapt to the corresponding human characteristics. Finally, we get $N_{feed}$ candidate lists after prepending each instruction into current adversarial instruction list separately. We use $N_{feed}$ as 10 and the number of pairs in $B_{tr}$ as 4 in our experiment.

\paragraph{Instruction Selection}
For each candidate, we collect generations on a validation batch of input content and human answer pairs $B_{val} = {(x_{val}^{1}, h_{val}^{1}), ..., (x_{val}^{|N_{val}|}, h_{val}^{|N_{val}|})}$ from the validation set $D_{val}$, separate from $D_{tr}$. We measure the scores of the target detector $f$ on them, and select top-k instruction lists that achieves the lowest accuracy. 

After selecting the top-k lists, we further optimize the expressions for each instruction through paraphrasing. We follow \citet{zhou2022large} to ask the LLM to generate a paraphrase for the newly added instruction in top-k lists. We collect $N_{para}$ paraphrases for each instruction. The final top-k candidates among the original top-k candidates and $N_{para}$ paraphrases are selected with the same process as above. We generate LLM responses with $D_{val}$, and choose k instruction lists that achieve the worst accuracy out of $f$. 
\section{Exploiting Prompt-Specific Shortcuts of AIGT Detectors}
\label{sec:attack}
In this section, we leverage adversarial instructions from \attackMethod\ to evaluate the reliance of an existing detector to prompt-specific features.

\begin{table*}[t]
\centering
\small
\setlength{\tabcolsep}{6pt} 
\renewcommand{\arraystretch}{1} 
\begin{tabular}{cl|ccc|ccc|ccc} 
\toprule
\multicolumn{2}{c|}{\multirow{2}{*}{}} & \multicolumn{3}{c|}{ChatGPT detector} & \multicolumn{3}{c|}{Perplexity} & \multicolumn{3}{c}{DetectGPT} \\
\multicolumn{2}{c|}{} & ELI5 & XSum & SQuAD & ELI5 & XSum & SQuAD & ELI5 & XSum & SQuAD \\ 
\midrule
\multirow{6}{*}{\begin{tabular}[c]{@{}c@{}}AUROC\\($\downarrow$)\end{tabular}} & N/A & 98.23 & 86.16 & 91.84 & 97.19 & 87.14 & 96.77 & 91.80 & 79.45 & 92.51 \\
 & PARA & 95.54 & 85.34 & 89.74 & 93.69 & 82.34 & 95.70 & 85.79 & 75.53 & 90.43 \\
 & DIPPER & 93.10 & 80.17 & 82.82 & 92.28 & 72.25 & 91.20 & 81.08 & \textbf{65.16} & 85.96 \\
 & SICO & 88.06 & 83.85 & \textbf{38.72} & 93.26 & 88.75 & 79.11 & 86.51 & 81.00 & \textbf{85.17} \\
 & IP & 94.07 & 72.82 & 88.07 & 96.24 & \textbf{68.86} & 94.90 & 90.72 & 73.77 & 92.17 \\
 & \attackMethod & \textbf{62.49} & \textbf{63.96} & 44.52 & \textbf{55.69} & 70.54 & \textbf{70.14} & \textbf{76.61} & 74.12 & 87.19 \\ 
\midrule
\multirow{5}{*}{\begin{tabular}[c]{@{}c@{}}ASR\\($\uparrow$)\end{tabular}} & PARA & 18.18 & 11.71 & 15.58 & 11.45 & 19.80 & 4.57 & 27.43 & 16.30 & 21.84 \\
 & DIPPER & 36.93 & 19.39 & 24.20 & 20.38 & 38.32 & 12.91 & 32.16 & \textbf{27.55} & 27.03 \\
 & SICO & 44.03 & 15.48 & 83.83 & 20.66 & 11.37 & 44.50 & 16.69 & 11.28 & \textbf{34.82} \\
 & IP & 17.29 & 33.85 & 19.53 & 4.62 & 47.36 & 8.12 & 15.45 & 19.61 & 14.57 \\
 & \attackMethod & \textbf{95.72} & \textbf{55.75} & \textbf{90.93} & \textbf{85.98} & \textbf{47.65} & \textbf{59.73} & \textbf{48.56} & 17.86 & 29.71 \\
\bottomrule
\end{tabular}
\caption{Detection performances on attacked ChatGPT (gpt-3.5-turbo-0613) generations. The N/A on AUROC represents the average AUROC of non-attack generations measured in the 5 attack methods. The best attack score for each column is represented in bold.}
\label{tab:attack_0613}
\end{table*}

\subsection{Setting}
\label{sec:attack_setting}
\paragraph{Datasets}
Two tasks are frequently used in AIGT detection: long-form question answering and text generation. 
We evaluate detectors on three English datasets from these tasks. 
For long-form question answering, we choose ELI5~\cite{eli5_lfqa}. For text generation, we choose XSum~\cite{narayan-etal-2018-dont} and SQuAD~\cite{rajpurkar2016squad}. 
More details are provided in Appendix~\ref{sec:appendix_dataset}.

\paragraph{AIGT Detectors}
We inspect the vulnerabilities relevant to prompt-specific shortcut features in ChatGPT detector \cite{guo-etal-2023-hc3}. ChatGPT detector is a RoBERTa-base \cite{liu2019roberta} detector fine-tuned to distinguish if a given text is written by human or ChatGPT~cite{chatgpt2023original}, trained on Human ChatGPT Comparison Corpus (HC3) \cite{guo-etal-2023-hc3}. HC3 consists of ChatGPT and human answers from five different tasks, where about 80\% are from ELI5 \cite{eli5_lfqa}.

We also assess the performance of two metric-based detectors, namely Perplexity \cite{jelinek1977perplexity} and DetectGPT \cite{mitchell2023detectgpt}, against the attack generations from ChatGPT detector experiment. Perplexity is based on the idea that the generation model will prefer AI texts to human ones. We measure the perplexity of a text from the proxy model and classify the texts with low perplexity as AI writing. DetectGPT detects AIGTs following the perturbation discrepancy gap hypothesis. Given a text, we perturb the text 100 times with T5-3b~\cite{roberts2019exploring} and compare the average probability of the perturbed outputs with the original output. If the probability decreases after perturbation, the original text is labeled as AI. If the probability does not change, the text is labeled as human.The original implementation often fails to perturb lengthy texts. Hence, we adopt the implementation of \citet{kirchenbauer2023reliability}.  We follow the default hyperparameters of \citet{mitchell2023detectgpt} in our experiment. 

The metric-based detectors require the probability of a text calculated by the generation model, which is not provided by the ChatGPT API. Therefore, we utilize another language model as a proxy. \citet{mireshghallah2023smaller} provides an extensive evaluation of various models for the DetectGPT method, reporting that OPT-125m~\cite{zhang2022opt} is the best universal detector, even when the generation model is much larger. Following this, OPT-125m serves as a proxy in our experiments.

\paragraph{Generation Model}
The train dataset of ChatGPT detector, HC3, is composed of generations from the early version of ChatGPT. To set the experiment setting close to the train setting of the detector, we utilize two versions of ChatGPT (gpt-3.5-turbo-0301, gpt-3.5-turbo-0613) in our experiments. 

\paragraph{Baseline Attacks}
Recent works found various attacks that perturb the output texts of LLMs to deceive AIGT detectors. We compare \attackMethod\ with several attacks to verify the significance of the vulnerability that \attackMethod\ exploits. 

\begin{itemize}[noitemsep,topsep=1pt]
    \item \textbf{N/A} generates texts from the base task description in Table \ref{tab:prompt_format_dataset} without any perturbation.
    \item \textbf{DIPPER} \cite{krishna2023paraphrasing} utilizes another model, DIPPER, to paraphrases the N/A generations. DIPPER is a variation of T5-XXL~\cite{roberts2019exploring} fine-tuned for paraphrasing.
    \item \textbf{PARA} refers to the self-paraphrases for the original responses. We simply ask the generation model to paraphrase its generations. 
    \item \textbf{SICO} \cite{lu2024large} iteratively searches for adversarial in-context examples that deceives AIGT detectors. First, LLM writes a description about a general difference between AI and human. Then, the model generates initial adversarial responses based on the description. SICO optimizes the examples to deceive detectors by alternating two substitution methods: WordNet-based word-level substitution and LLM-based sentence-level substitution. 
    \item \textbf{IP} \cite{shi2023red} also utilizes adversarial in-context examples to deceive detectors. It alternately generates candidates for the in-context example and the instruction asking to follow the example. The pair with the lowest detection score is selected to optimize its adversarial effect. 
\end{itemize}

We follow the original generation configuration for each attack. As each attack differs in the base prompt for each dataset and the length of generations, we modify the original prompts to match our experiment setting. We provide details of our implementations in Appendix \ref{sec:appendix_attack_methods}.

\paragraph{Details for \attackMethod} We iterate 6 times, and select top-2 instruction lists for each step. We select the instruction list with the lowest validation score as the final \attackMethod\ instruction list. When the model generates paraphrase instructions or responses corresponding to the generation task, we set the temperature as 1. For other steps, i.e. feedback generation and feedback conversion, we set the temperature as 0 to better reflect the assessment of the model.

\paragraph{Evaluation}

In our experiment, each attack is evaluated with 200 inputs from each dataset whose non-attack generation, attacked generation, and human answers are between 256 and 450 tokens. For each question, we truncate the three responses to match the length of to the shortest. This leads to slight differences in the non-attack generations and human answers among test results. To assure the validity of the comparison between test results, we also report the AUROC scores for non-attack generations on each test in Table~\ref{tab:original_scores}. We find the intra-task variance to be small.

\paragraph{Metrics} We evaluate detectors with AUROC and Attack Success Rate (ASR). ASR is calculated as the ratio of the number of inputs whose generations were originally detected, but not detected after attack, to the number of generations originally detected. As Perplexity and DetectGPT do not have pre-defined thresholds for classification, we set the detection threshold as the value that achieves the best F1 on N/A to measure ASR.

\subsection{Experiment Results}
Table \ref{tab:attack_0301} and \ref{tab:attack_0613} show the performance of AIGT detectors on generations from the two versions of ChatGPT. We test each setting on 3 random seeds and report the average values.
High AUROC scores on N/A show that the ChatGPT detector can easily discriminate generations from the base prompt. However, its performance is not resilient to attacks. The impact of \attackMethod\ is comparable to other baselines. Other detectors are also affected, but their drop is inconsistent and less than the drop of ChatGPT detector. On gpt-3.5-turbo-0301, the deceptive effect of \attackMethod\ generations does not generalize to others. \attackMethod\ generations from gpt-3.5-turbo-0613 significantly reduce detection scores of ChatGPT detector and Perplexity, but they are less effective on DetectGPT. This shows that the features perturbed by \attackMethod\ instructions do not represent the general behavior of the generation model, but ChatGPT detector shows high dependency towards such features, compared to metric-based detectors.

\section{Improving Robustness with \attackMethod\ Generations}
\label{sec:defense}

In Section~\ref{sec:attack}, we could exploit the overreliance of the ChatGPT detector on prompt-specific features to deceive them. If the failure is due to shortcut learning, augmenting train data with AIGTs from other prompts can improve its robustness as the additional data alleviates the dataset bias. 
In this section, we enhance the robustness of detectors against prompt variation through train data augmentation. A major challenge in this approach is finding prompts that effectively perturb major shortcut features. Since \attackMethod\ proved its effectiveness in finding such instructions, we leverage instructions from a \attackMethod\ run for augmentation.

\begin{table}[t]
\centering
\small
\setlength{\tabcolsep}{2.5pt} 
\renewcommand{\arraystretch}{1} 
\begin{adjustbox}{width=1\linewidth}
\begin{tabular}{lcccccc} 
\toprule
\multirow{2}{*}{} & \multicolumn{3}{c}{ChatGPT detector} & \multicolumn{3}{c}{Augmented} \\
 & ELI5 & XSum & SQuAD & ELI5 & XSum & SQuAD \\ 
\midrule
\multicolumn{7}{c}{gpt-3.5-turbo-0301} \\ 
\midrule
N/A & 93.33 & 80.70 & 96.08 & {\cellcolor[rgb]{0.9,0.9,0.9}}100.00 & {\cellcolor[rgb]{0.9,0.9,0.9}}98.07 & {\cellcolor[rgb]{0.9,0.9,0.9}}99.01 \\ 
PARA & 89.31 & 63.61 & 83.13 & {\cellcolor[rgb]{0.9,0.9,0.9}}100.00 & {\cellcolor[rgb]{0.9,0.9,0.9}}98.99 & {\cellcolor[rgb]{0.9,0.9,0.9}}99.10 \\ 
DIPPER & 92.27 & 80.71 & 84.84 & {\cellcolor[rgb]{0.9,0.9,0.9}}99.44 & {\cellcolor[rgb]{0.9,0.9,0.9}}88.30 & {\cellcolor[rgb]{0.9,0.9,0.9}}85.42 \\ 
SICO & 77.21 & 50.31 & 61.45 & {\cellcolor[rgb]{0.9,0.9,0.9}}99.93 & {\cellcolor[rgb]{0.9,0.9,0.9}}95.87 & {\cellcolor[rgb]{0.9,0.9,0.9}}98.97 \\ 
IP & 89.63 & 40.47 & 75.51 & {\cellcolor[rgb]{0.9,0.9,0.9}}100.00 & {\cellcolor[rgb]{0.9,0.9,0.9}}88.20 & {\cellcolor[rgb]{0.9,0.9,0.9}}98.67 \\ 
FAILOpt & 78.17 & 64.92 & 88.31 & {\cellcolor[rgb]{0.9,0.9,0.9}}100.00 & {\cellcolor[rgb]{0.9,0.9,0.9}}90.76 & {\cellcolor[rgb]{0.9,0.9,0.9}}98.87 \\
\midrule
\multicolumn{7}{c}{gpt-3.5-turbo-0613} \\ 
\midrule
N/A & 98.23 & 86.16 & 91.84 & {\cellcolor[rgb]{0.9,0.9,0.9}}100.00 & {\cellcolor[rgb]{0.9,0.9,0.9}}98.91 & {\cellcolor[rgb]{0.9,0.9,0.9}}98.98 \\
PARA & 95.54 & 85.34 & 89.74 & {\cellcolor[rgb]{0.9,0.9,0.9}}100.00 & {\cellcolor[rgb]{0.9,0.9,0.9}}98.63 & {\cellcolor[rgb]{0.9,0.9,0.9}}98.87 \\
DIPPER & 93.10 & 80.17 & 82.82 & {\cellcolor[rgb]{0.9,0.9,0.9}}99.72 & {\cellcolor[rgb]{0.9,0.9,0.9}}94.81 & {\cellcolor[rgb]{0.9,0.9,0.9}}90.24 \\
SICO & 88.06 & 83.85 & 38.72 & {\cellcolor[rgb]{0.9,0.9,0.9}}99.99 & {\cellcolor[rgb]{0.9,0.9,0.9}}98.12 & {\cellcolor[rgb]{0.9,0.9,0.9}}98.80 \\
IP & 94.07 & 72.82 & 88.07 & {\cellcolor[rgb]{0.9,0.9,0.9}}100.00 & {\cellcolor[rgb]{0.9,0.9,0.9}}98.78 & {\cellcolor[rgb]{0.9,0.9,0.9}}98.74 \\
FAILOpt & 62.49 & 63.96 & 44.52 & {\cellcolor[rgb]{0.9,0.9,0.9}}100.00 & {\cellcolor[rgb]{0.9,0.9,0.9}}98.99 & {\cellcolor[rgb]{0.9,0.9,0.9}}98.88 \\
\bottomrule
\end{tabular}
\end{adjustbox}
\caption{AUROC of the original and re-trained detectors on generations of ChatGPTs (gpt-3.5-turbo-0301, gpt-3.5-turbo-0613) in percentage. The detector enhances in every setting after training on the augmented data.}
\label{tab:robust_main_all_auroc}
\end{table}

\subsection{Augmentation Setting}

\paragraph{Data Collection}
To minimize the influence of domain difference, we construct a binary classification dataset from ELI5, which accounts for a major portion of HC3. We select 2000 ELI5 questions not included in HC3. Then, for each question, we gather a human answer, an AIGT from the base prompt, and an AIGT from a \attackMethod\ prompt. Following \citet{guo-etal-2023-hc3},  each sentence in the full answers is also utilized as a training sample. We split each sentence from the full answers with NLTK \cite{bird2009natural} library. Generations from both prompts are labeled as 'ChatGPT'. We used the following instructions found in a single \attackMethod\ run on ELI5 for data augmentation:

\begin{tcolorbox}[title = {{\attackMethod\ Instructions For Augmentation}}]
Incorporate witty remarks and irony to convey your message in your responses.

Please provide structured and organized answers.

Incorporate detailed instances and jargon into your responses.

Incorporate humor or sarcasm into your responses.
\end{tcolorbox}

\paragraph{Train setting}
We re-train the ChatGPT detector on our dataset using 5 random seeds and follow the hyperparameters in \citet{guo-etal-2023-hc3} for training. Each training takes less than an hour on two 16GB NVIDIA V100 gpus.

\subsection{Robustness Evaluation}
Table~\ref{tab:robust_main_all_auroc} compares the average AUROC of five augmented detectors to the original ChatGPT detector in each dataset. We find that the augmentation significantly enhances the detection performance in every setting, regardless of dataset, generation method, and version of ChatGPT. Also, despite the train data shift, our detectors do not suffer from the trade-off between N/A and attacked generations. This result supports that the detectors effectively learn the general features of generations via our data augmentation approach.

\begin{table}[t]
\centering
\small
\setlength{\tabcolsep}{3pt} 
\renewcommand{\arraystretch}{1} 
\begin{tabular}{clcccc} 
\toprule
\multirow{2}{*}{Model} & \multirow{2}{*}{Attack} & \multicolumn{4}{c}{Train Data Sources} \\
 &  & \multicolumn{1}{l}{No train} & Full & - N/A & - \attackMethod \\ 
\midrule
\multirow{6}{*}{0301} & N/A & 9.50 & 0.71 & 11.89 & 1.65 \\
 & PARA & 23.49 & 0.51 & 11.85 & 0.96 \\
 & DIPPER & 17.11 & 10.41 & 34.85 & 14.58 \\
 & SICO & 43.06 & 2.12 & 13.39 & 25.80 \\
 & IP & 34.52 & 2.68 & 18.56 & 4.62 \\
 & \attackMethod & 28.36 & 2.04 & 10.02 & 26.09 \\ 
\midrule
\multirow{6}{*}{0613} & N/A & 4.83 & 0.43 & 9.44 & 0.86 \\
 & PARA & 6.38 & 0.43 & 9.20 & 0.82 \\
 & DIPPER & 17.79 & 6.36 & 27.26 & 10.01 \\
 & SICO & 36.88 & 0.70 & 9.20 & 30.65 \\
 & IP & 12.75 & 0.49 & 9.46 & 1.04 \\
 & \attackMethod & 57.68 & 0.44 & 9.13 & 8.73 \\
\bottomrule
\end{tabular}
\caption{Average human score of AIGTs from the test datasets in percentage. Full achieves the best score against every generation method.}
\label{tab:robust_main_0301_auroc}
\end{table}
\subsection{Discussion}
We compare the impact of training detectors on 2000 texts from various data sources. Evaluation is conducted on four settings. \textbf{No train} refers to the original ChatGPT detector without additional training. \textbf{Full} represents detectors trained on data from all sources. (i. e., human, N/A, and \attackMethod\ generations). \textbf{- N/A} represents detectors trained with only human and \attackMethod\ generations. \textbf{- \attackMethod} represents detectors trained with only human and N/A generations. 

We find that detectors trained on data from different prompts learn different features. \textbf{- \attackMethod} is weak against SICO and \attackMethod\ outputs, and \textbf{- N/A} is weak against N/A, DIPPER, and PARA. In contrast, \textbf{Full} achieves a high score in all cases, although the generated data are shared by either \textbf{- \attackMethod} or \textbf{- N/A}. This result implies that generations from the \attackMethod\ prompt provide data complementary to the base prompt generations. Each of the prompts biases the model differently, but the \attackMethod\ prompt generations are biased in a way that conflicts with major shortcut features in the base prompt generations. Hence, \textbf{Full} learns general features that do not rely on shortcuts that the original ChatGPT detector relied upon.

As we re-train a fully trained detector, the prior of the detector can affect the train result, especially in the early stage. Therefore, we further inspect the impact of train data with the change of human score, the likelihood to be human writing measured by the detector, on AIGTs as the number of train data increases from 500 to 2000. Human score of text $g$ is measured as
$hs(g) = 1 - f(g)$.

\begin{figure}[t]
\includegraphics[width=\linewidth]{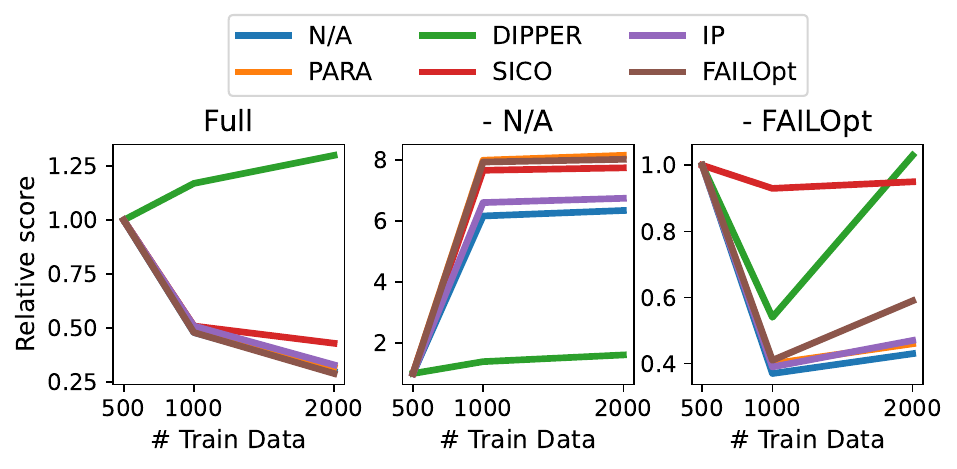}
\caption{The change of human scores on various attack generations from gpt-3.5-turbo-0613 as the number of train data increases. Except for DIPPER, the scores monotonically decrease only in \textbf{Full}.}
\label{fig:ablation_data_number}
\end{figure}

Figure \ref{fig:ablation_data_number} shows the ratio of average human scores of 0613 generations from the three datasets between each train data size and 500. As the number of data increases, \textbf{Full} generally improves in every dataset against every input perturbation attack, i.e. SICO, IP, and \attackMethod. \textbf{- \attackMethod} and \textbf{- N/A} does not follow this observation. In terms of the human scores of AI generations, \textbf{- N/A} reaches the lowest score at 500, but quickly degrades surpassing 1000. \textbf{- \attackMethod} also deteriorates when train data from each source increases from 1000 to 2000. This result also confirms undesirable biases in data from a single prompt, but augmentation with \attackMethod\ instructions alleviate the issue. One exception in the ablation is DIPPER: as the train data gets larger, even \textbf{Full} slowly loses its robustness to DIPPER generations. We posit that this weakness stems from the model shift. Unlike other generation methods, DIPPER leverages another model for text perturbaton. As aforementioned general features are still bound to the data collection model we used, the performance on other models can worsen. Note that \textbf{Full} still achieves the best score against DIPPER. See Appendix \ref{sec:augmented_full} for full result.

\section{Conclusion}
\label{sec:conclusion}
We show that AIGT detectors trained on data generated with limited prompts can be unreliable as it is susceptible to learning prompt-specific shortcuts.
To this end, we first verify that there are instructions that elude detectors by negating the prompt-specific behavior of an LLM. We propose \attackMethodFull, an attack that exploits prompt-specific shortcuts to find instructions that elude detectors effectively.
Then, we utilize a \attackMethod\ prompt to train a reliable detector. Re-training the vulnerable detector generally improves on various datasets and generation methods. This implies that preventing shortcut learning plays a key role in the development of reliable AIGT detectors, and \attackMethod\ can effectively mitigate shortcuts.

\section*{Limitations}
We introduce a simple method to improve the robustness of detectors via data augmentation. However, other sources of non-robust features remain not covered in our approach. For example, the ablation results show that the improvement is limited against generations perturbed with another model. To develop a detector robust to changes in any generation settings, we should construct a comprehensive dataset that includes other types of variations. Our work concentrate on showing the importance of prompt variation, an important factor frequently overlooked in previous literature. We leave the construction of the comprehensive dataset as future work.

Also, we do not suggest a method to improve metric-based detectors in this paper. Unlike the supervised classifiers, we cannot adjust metric-based detectors with additional data. Instead, we should come up with a novel metric that illustrates characteristics of LLMs that are consistent and irrelevant to prompt choices. This is an important topic for the development of a reliable zero-shot AIGT detector, and we expect future studies.

\section*{Ethical Considerations}
While navigating the issue of prompt-specific shortcuts, we reveal weaknesses of existing AIGT detectors. We do not intend to encourage abusive uses with \attackMethod. Instead, we spotlight an important topic that was overlooked in previous works: the importance of diverse data collection prompts in AIGT detection. The proposed attack, \attackMethod, is provided as a tool to measure the influence of prompt-specific shortcuts and raise concern about this issue to the researcher community. Also, we offer a simple, easily applicable defense against input perturbation attacks leveraging \attackMethod. We hope the suggested defense approach prevents the malignant uses of LLMs, and contributes to the development of a reliable AIGT detector.   


\begin{thebibliography}{48}
\expandafter\ifx\csname natexlab\endcsname\relax\def\natexlab#1{#1}\fi

\bibitem[{Achiam et~al.(2023)Achiam, Adler, Agarwal, Ahmad, Akkaya, Aleman, Almeida, Altenschmidt, Altman, Anadkat et~al.}]{achiam2023gpt}
Josh Achiam, Steven Adler, Sandhini Agarwal, Lama Ahmad, Ilge Akkaya, Florencia~Leoni Aleman, Diogo Almeida, Janko Altenschmidt, Sam Altman, Shyamal Anadkat, et~al. 2023.
\newblock Gpt-4 technical report.
\newblock \emph{arXiv preprint arXiv:2303.08774}.

\bibitem[{Anthropic(2024)}]{anthropic2024claude}
AI~Anthropic. 2024.
\newblock The claude 3 model family: Opus, sonnet, haiku.
\newblock \emph{Claude-3 Model Card}.

\bibitem[{Bao et~al.(2023)Bao, Zhao, Teng, Yang, and Zhang}]{bao2023fast}
Guangsheng Bao, Yanbin Zhao, Zhiyang Teng, Linyi Yang, and Yue Zhang. 2023.
\newblock Fast-detectgpt: Efficient zero-shot detection of machine-generated text via conditional probability curvature.
\newblock In \emph{The Twelfth International Conference on Learning Representations}.

\bibitem[{Bird et~al.(2009)Bird, Klein, and Loper}]{bird2009natural}
Steven Bird, Ewan Klein, and Edward Loper. 2009.
\newblock \emph{Natural language processing with Python: analyzing text with the natural language toolkit}.
\newblock " O'Reilly Media, Inc.".

\bibitem[{Bohacek(2023)}]{bohacek2023unseen}
Matyas Bohacek. 2023.
\newblock \href {https://openreview.net/pdf?id=9ZKJLYg5EQ} {{The Unseen A+ Student: Navigating the Impact of Large Language Models in the Classroom}}.
\newblock In \emph{ICML 2023 Workshop on Deployment Challenges for Generative AI}.

\bibitem[{Brown et~al.(2020)Brown, Mann, Ryder, Subbiah, Kaplan, Dhariwal, Neelakantan, Shyam, Sastry, Askell et~al.}]{brown2020language}
Tom Brown, Benjamin Mann, Nick Ryder, Melanie Subbiah, Jared~D Kaplan, Prafulla Dhariwal, Arvind Neelakantan, Pranav Shyam, Girish Sastry, Amanda Askell, et~al. 2020.
\newblock Language models are few-shot learners.
\newblock \emph{Advances in neural information processing systems}, 33:1877--1901.

\bibitem[{Busch and Hausvik(2023)}]{busch2023too}
Peter~André Busch and Geir~Inge Hausvik. 2023.
\newblock Too good to be true? an empirical study of chatgpt capabilities for academic writing and implications for academic misconduct.
\newblock In \emph{AMCIS 2023 Proceedings}.

\bibitem[{Chen et~al.(2023)Chen, Kang, Zhai, Li, Singh, and Ramakrishnan}]{chen2023gpt}
Yutian Chen, Hao Kang, Vivian Zhai, Liangze Li, Rita Singh, and Bhiksha Ramakrishnan. 2023.
\newblock Gpt-sentinel: Distinguishing human and chatgpt generated content.
\newblock \emph{arXiv preprint arXiv:2305.07969}.

\bibitem[{Du et~al.(2023)Du, He, Zou, Tao, and Hu}]{du2023shortcut}
Mengnan Du, Fengxiang He, Na~Zou, Dacheng Tao, and Xia Hu. 2023.
\newblock Shortcut learning of large language models in natural language understanding.
\newblock \emph{Communications of the ACM}, 67(1):110--120.

\bibitem[{Fan et~al.(2019)Fan, Jernite, Perez, Grangier, Weston, and Auli}]{eli5_lfqa}
Angela Fan, Yacine Jernite, Ethan Perez, David Grangier, Jason Weston, and Michael Auli. 2019.
\newblock \href {https://doi.org/10.18653/v1/p19-1346} {{ELI5:} long form question answering}.
\newblock In \emph{Proceedings of the 57th Conference of the Association for Computational Linguistics, {ACL} 2019, Florence, Italy, July 28- August 2, 2019, Volume 1: Long Papers}, pages 3558--3567. Association for Computational Linguistics.

\bibitem[{Fernando et~al.(2023)Fernando, Banarse, Michalewski, Osindero, and Rockt{\"a}schel}]{fernando2023promptbreeder}
Chrisantha Fernando, Dylan Banarse, Henryk Michalewski, Simon Osindero, and Tim Rockt{\"a}schel. 2023.
\newblock Promptbreeder: Self-referential self-improvement via prompt evolution.
\newblock \emph{arXiv preprint arXiv:2309.16797}.

\bibitem[{Geirhos et~al.(2020)Geirhos, Jacobsen, Michaelis, Zemel, Brendel, Bethge, and Wichmann}]{geirhos2020shortcut}
Robert Geirhos, J{\"o}rn-Henrik Jacobsen, Claudio Michaelis, Richard Zemel, Wieland Brendel, Matthias Bethge, and Felix~A Wichmann. 2020.
\newblock Shortcut learning in deep neural networks.
\newblock \emph{Nature Machine Intelligence}, 2(11):665--673.

\bibitem[{Guo et~al.(2023)Guo, Zhang, Wang, Jiang, Nie, Ding, Yue, and Wu}]{guo-etal-2023-hc3}
Biyang Guo, Xin Zhang, Ziyuan Wang, Minqi Jiang, Jinran Nie, Yuxuan Ding, Jianwei Yue, and Yupeng Wu. 2023.
\newblock How close is chatgpt to human experts? comparison corpus, evaluation, and detection.
\newblock \emph{arXiv preprint arxiv:2301.07597}.

\bibitem[{Ha et~al.(2023)Ha, Tran, and Kim}]{ha2023black}
Huyen Ha, Duc Tran, and Dukyun Kim. 2023.
\newblock Black-box adversarial attacks against language model detector.
\newblock In \emph{Proceedings of the 12th International Symposium on Information and Communication Technology}, pages 754--760.

\bibitem[{Hans et~al.(2024)Hans, Schwarzschild, Cherepanova, Kazemi, Saha, Goldblum, Geiping, and Goldstein}]{hans2024spotting}
Abhimanyu Hans, Avi Schwarzschild, Valeriia Cherepanova, Hamid Kazemi, Aniruddha Saha, Micah Goldblum, Jonas Geiping, and Tom Goldstein. 2024.
\newblock \href {http://arxiv.org/abs/2401.12070} {Spotting llms with binoculars: Zero-shot detection of machine-generated text}.

\bibitem[{Hermann et~al.(2024)Hermann, Mobahi, FEL, and Mozer}]{hermann2024on}
Katherine Hermann, Hossein Mobahi, Thomas FEL, and Michael~Curtis Mozer. 2024.
\newblock \href {https://openreview.net/forum?id=Tj3xLVuE9f} {On the foundations of shortcut learning}.
\newblock In \emph{The Twelfth International Conference on Learning Representations}.

\bibitem[{Huang et~al.(2024)Huang, Zhang, Li, You, Wang, and Yang}]{huang2024ai}
Guanhua Huang, Yuchen Zhang, Zhe Li, Yongjian You, Mingze Wang, and Zhouwang Yang. 2024.
\newblock Are ai-generated text detectors robust to adversarial perturbations?
\newblock \emph{arXiv preprint arXiv:2406.01179}.

\bibitem[{Jelinek et~al.(1977)Jelinek, Mercer, Bahl, and Baker}]{jelinek1977perplexity}
Fred Jelinek, Robert~L Mercer, Lalit~R Bahl, and James~K Baker. 1977.
\newblock Perplexity—a measure of the difficulty of speech recognition tasks.
\newblock \emph{The Journal of the Acoustical Society of America}, 62(S1):S63--S63.

\bibitem[{Kirchenbauer et~al.(2023{\natexlab{a}})Kirchenbauer, Geiping, Wen, Katz, Miers, and Goldstein}]{kirchenbauer2023watermark}
John Kirchenbauer, Jonas Geiping, Yuxin Wen, Jonathan Katz, Ian Miers, and Tom Goldstein. 2023{\natexlab{a}}.
\newblock \href {http://arxiv.org/abs/2301.10226} {A watermark for large language models}.

\bibitem[{Kirchenbauer et~al.(2023{\natexlab{b}})Kirchenbauer, Geiping, Wen, Shu, Saifullah, Kong, Fernando, Saha, Goldblum, and Goldstein}]{kirchenbauer2023reliability}
John Kirchenbauer, Jonas Geiping, Yuxin Wen, Manli Shu, Khalid Saifullah, Kezhi Kong, Kasun Fernando, Aniruddha Saha, Micah Goldblum, and Tom Goldstein. 2023{\natexlab{b}}.
\newblock \href {http://arxiv.org/abs/2306.04634} {On the reliability of watermarks for large language models}.

\bibitem[{Koike et~al.(2023)Koike, Kaneko, and Okazaki}]{koike2023you}
Ryuto Koike, Masahiro Kaneko, and Naoaki Okazaki. 2023.
\newblock How you prompt matters! even task-oriented constraints in instructions affect llm-generated text detection.
\newblock \emph{arXiv preprint arXiv:2311.08369}.

\bibitem[{Koike et~al.(2024)Koike, Kaneko, and Okazaki}]{koike2024outfox}
Ryuto Koike, Masahiro Kaneko, and Naoaki Okazaki. 2024.
\newblock Outfox: Llm-generated essay detection through in-context learning with adversarially generated examples.
\newblock In \emph{Proceedings of the 38th AAAI Conference on Artificial Intelligence}, Vancouver, Canada.

\bibitem[{Krishna et~al.(2023)Krishna, Song, Karpinska, Wieting, and Iyyer}]{krishna2023paraphrasing}
Kalpesh Krishna, Yixiao Song, Marzena Karpinska, John Wieting, and Mohit Iyyer. 2023.
\newblock Paraphrasing evades detectors of ai-generated text, but retrieval is an effective defense.
\newblock \emph{arXiv preprint arXiv:2303.13408}.

\bibitem[{Kuditipudi et~al.(2023)Kuditipudi, Thickstun, Hashimoto, and Liang}]{kuditipudi2023robust}
Rohith Kuditipudi, John Thickstun, Tatsunori Hashimoto, and Percy Liang. 2023.
\newblock Robust distortion-free watermarks for language models.
\newblock \emph{arXiv preprint arXiv:2307.15593}.

\bibitem[{Li et~al.(2023)Li, Wang, Zhang, Zhu, Hou, Lian, Luo, Yang, and Xie}]{li2023large}
Cheng Li, Jindong Wang, Yixuan Zhang, Kaijie Zhu, Wenxin Hou, Jianxun Lian, Fang Luo, Qiang Yang, and Xing Xie. 2023.
\newblock Large language models understand and can be enhanced by emotional stimuli.
\newblock \emph{arXiv preprint arXiv:2307.11760}.

\bibitem[{Liu et~al.(2019)Liu, Ott, Goyal, Du, Joshi, Chen, Levy, Lewis, Zettlemoyer, and Stoyanov}]{liu2019roberta}
Yinhan Liu, Myle Ott, Naman Goyal, Jingfei Du, Mandar Joshi, Danqi Chen, Omer Levy, Mike Lewis, Luke Zettlemoyer, and Veselin Stoyanov. 2019.
\newblock Roberta: A robustly optimized bert pretraining approach.
\newblock \emph{arXiv preprint arXiv:1907.11692}.

\bibitem[{Lu et~al.(2024)Lu, Liu, He, Ong, Wang, and Tang}]{lu2024large}
Ning Lu, Shengcai Liu, Rui He, Yew-Soon Ong, Qi~Wang, and Ke~Tang. 2024.
\newblock \href {https://openreview.net/forum?id=lLE0mWzUrr} {Large language models can be guided to evade {AI}-generated text detection}.
\newblock \emph{Transactions on Machine Learning Research}.

\bibitem[{Lu et~al.(2022)Lu, Bartolo, Moore, Riedel, and Stenetorp}]{lu-etal-2022-fantastically}
Yao Lu, Max Bartolo, Alastair Moore, Sebastian Riedel, and Pontus Stenetorp. 2022.
\newblock \href {https://doi.org/10.18653/v1/2022.acl-long.556} {Fantastically ordered prompts and where to find them: Overcoming few-shot prompt order sensitivity}.
\newblock In \emph{Proceedings of the 60th Annual Meeting of the Association for Computational Linguistics (Volume 1: Long Papers)}, pages 8086--8098, Dublin, Ireland. Association for Computational Linguistics.

\bibitem[{Mireshghallah et~al.(2023)Mireshghallah, Mattern, Gao, Shokri, and Berg-Kirkpatrick}]{mireshghallah2023smaller}
Fatemehsadat Mireshghallah, Justus Mattern, Sicun Gao, Reza Shokri, and Taylor Berg-Kirkpatrick. 2023.
\newblock \href {http://arxiv.org/abs/2305.09859} {Smaller language models are better black-box machine-generated text detectors}.

\bibitem[{Mitchell et~al.(2023)Mitchell, Lee, Khazatsky, Manning, and Finn}]{mitchell2023detectgpt}
Eric Mitchell, Yoonho Lee, Alexander Khazatsky, Christopher~D. Manning, and Chelsea Finn. 2023.
\newblock \href {https://arxiv.org/abs/2301.11305} {Detectgpt: Zero-shot machine-generated text detection using probability curvature}.

\bibitem[{Montani et~al.(2023)Montani, Honnibal, Honnibal, Boyd, Van~Landeghem, and Peters}]{montani2023spacy}
Ines Montani, Matthew Honnibal, Matthew Honnibal, Adriane Boyd, Sofie Van~Landeghem, and Henning Peters. 2023.
\newblock \href {https://doi.org/10.5281/ZENODO.1212303} {explosion/spacy: v3.7.2: Fixes for apis and requirements}.

\bibitem[{Narayan et~al.(2018)Narayan, Cohen, and Lapata}]{narayan-etal-2018-dont}
Shashi Narayan, Shay~B. Cohen, and Mirella Lapata. 2018.
\newblock \href {https://doi.org/10.18653/v1/D18-1206} {Don{'}t give me the details, just the summary! topic-aware convolutional neural networks for extreme summarization}.
\newblock In \emph{Proceedings of the 2018 Conference on Empirical Methods in Natural Language Processing}, pages 1797--1807, Brussels, Belgium. Association for Computational Linguistics.

\bibitem[{Pan et~al.(2023)Pan, Pan, Chen, Nakov, Kan, and Wang}]{pan2023risk}
Yikang Pan, Liangming Pan, Wenhu Chen, Preslav Nakov, Min-Yen Kan, and William~Yang Wang. 2023.
\newblock \href {http://arxiv.org/abs/2305.13661} {On the risk of misinformation pollution with large language models}.

\bibitem[{Raffel et~al.(2020)Raffel, Shazeer, Roberts, Lee, Narang, Matena, Zhou, Li, and Liu}]{roberts2019exploring}
Colin Raffel, Noam Shazeer, Adam Roberts, Katherine Lee, Sharan Narang, Michael Matena, Yanqi Zhou, Wei Li, and Peter~J. Liu. 2020.
\newblock \href {http://jmlr.org/papers/v21/20-074.html} {Exploring the limits of transfer learning with a unified text-to-text transformer}.
\newblock \emph{Journal of Machine Learning Research}, 21(140):1--67.

\bibitem[{Rajpurkar et~al.(2016)Rajpurkar, Zhang, Lopyrev, and Liang}]{rajpurkar2016squad}
Pranav Rajpurkar, Jian Zhang, Konstantin Lopyrev, and Percy Liang. 2016.
\newblock Squad: 100,000+ questions for machine comprehension of text.
\newblock In \emph{Proceedings of the 2016 Conference on Empirical Methods in Natural Language Processing}, pages 2383--2392.

\bibitem[{Sadasivan et~al.(2023)Sadasivan, Kumar, Balasubramanian, Wang, and Feizi}]{sadasivan2023aigenerated}
Vinu~Sankar Sadasivan, Aounon Kumar, Sriram Balasubramanian, Wenxiao Wang, and Soheil Feizi. 2023.
\newblock \href {http://arxiv.org/abs/2303.11156} {Can ai-generated text be reliably detected?}

\bibitem[{Sahoo et~al.(2024)Sahoo, Singh, Saha, Jain, Mondal, and Chadha}]{sahoo2024systematic}
Pranab Sahoo, Ayush~Kumar Singh, Sriparna Saha, Vinija Jain, Samrat Mondal, and Aman Chadha. 2024.
\newblock A systematic survey of prompt engineering in large language models: Techniques and applications.
\newblock \emph{arXiv preprint arXiv:2402.07927}.

\bibitem[{Sclar et~al.(2023)Sclar, Choi, Tsvetkov, and Suhr}]{sclar2023quantifying}
Melanie Sclar, Yejin Choi, Yulia Tsvetkov, and Alane Suhr. 2023.
\newblock Quantifying language models' sensitivity to spurious features in prompt design or: How i learned to start worrying about prompt formatting.
\newblock \emph{arXiv preprint arXiv:2310.11324}.

\bibitem[{Shi et~al.(2024)Shi, Wang, Yin, Chen, Chang, and Hsieh}]{shi2023red}
Zhouxing Shi, Yihan Wang, Fan Yin, Xiangning Chen, Kai-Wei Chang, and Cho-Jui Hsieh. 2024.
\newblock Red teaming language model detectors with language models.
\newblock \emph{Transactions of the Association for Computational Linguistics}, 12:174--189.

\bibitem[{Spitale et~al.(2023)Spitale, Biller-Andorno, and Germani}]{spitale2023ai}
Giovanni Spitale, Nikola Biller-Andorno, and Federico Germani. 2023.
\newblock Ai model gpt-3 (dis) informs us better than humans.
\newblock \emph{Science Advances}, 9(26):eadh1850.

\bibitem[{Su et~al.(2023{\natexlab{a}})Su, Zhuo, Wang, and Nakov}]{su2023detectllm}
Jinyan Su, Terry~Yue Zhuo, Di~Wang, and Preslav Nakov. 2023{\natexlab{a}}.
\newblock \href {http://arxiv.org/abs/2306.05540} {Detectllm: Leveraging log rank information for zero-shot detection of machine-generated text}.

\bibitem[{Su et~al.(2023{\natexlab{b}})Su, Wu, Zhou, Ma, and Hu}]{su2023hc3}
Zhenpeng Su, Xing Wu, Wei Zhou, Guangyuan Ma, and Songlin Hu. 2023{\natexlab{b}}.
\newblock Hc3 plus: A semantic-invariant human chatgpt comparison corpus.
\newblock \emph{arXiv preprint arXiv:2309.02731}.

\bibitem[{Taguchi et~al.(2024)Taguchi, Gu, and Sakurai}]{taguchi2024impact}
Kaito Taguchi, Yujie Gu, and Kouichi Sakurai. 2024.
\newblock The impact of prompts on zero-shot detection of ai-generated text.
\newblock \emph{arXiv preprint arXiv:2403.20127}.

\bibitem[{Touvron et~al.(2023)Touvron, Martin, Stone, Albert, Almahairi, Babaei, Bashlykov, Batra, Bhargava, Bhosale et~al.}]{touvron2023llama}
Hugo Touvron, Louis Martin, Kevin Stone, Peter Albert, Amjad Almahairi, Yasmine Babaei, Nikolay Bashlykov, Soumya Batra, Prajjwal Bhargava, Shruti Bhosale, et~al. 2023.
\newblock Llama 2: Open foundation and fine-tuned chat models.
\newblock \emph{arXiv preprint arXiv:2307.09288}.

\bibitem[{Tulchinskii et~al.(2023)Tulchinskii, Kuznetsov, Laida, Cherniavskii, Nikolenko, Burnaev, Barannikov, and Piontkovskaya}]{tulchinskii2023intrinsic}
Eduard Tulchinskii, Kristian Kuznetsov, Kushnareva Laida, Daniil Cherniavskii, Sergey Nikolenko, Evgeny Burnaev, Serguei Barannikov, and Irina Piontkovskaya. 2023.
\newblock \href {https://openreview.net/forum?id=8uOZ0kNji6} {Intrinsic dimension estimation for robust detection of {AI}-generated texts}.
\newblock In \emph{Thirty-seventh Conference on Neural Information Processing Systems}.

\bibitem[{Zhang et~al.(2022)Zhang, Roller, Goyal, Artetxe, Chen, Chen, Dewan, Diab, Li, Lin, Mihaylov, Ott, Shleifer, Shuster, Simig, Koura, Sridhar, Wang, and Zettlemoyer}]{zhang2022opt}
Susan Zhang, Stephen Roller, Naman Goyal, Mikel Artetxe, Moya Chen, Shuohui Chen, Christopher Dewan, Mona Diab, Xian Li, Xi~Victoria Lin, Todor Mihaylov, Myle Ott, Sam Shleifer, Kurt Shuster, Daniel Simig, Punit~Singh Koura, Anjali Sridhar, Tianlu Wang, and Luke Zettlemoyer. 2022.
\newblock \href {http://arxiv.org/abs/2205.01068} {Opt: Open pre-trained transformer language models}.

\bibitem[{Zhang et~al.(2023)Zhang, Zhang, Wang, and Jin}]{zhang2023assaying}
Yi-Fan Zhang, Zhang Zhang, Liang Wang, and Rong Jin. 2023.
\newblock \href {http://arxiv.org/abs/2312.12918} {Assaying on the robustness of zero-shot machine-generated text detectors}.

\bibitem[{Zhou et~al.(2022)Zhou, Muresanu, Han, Paster, Pitis, Chan, and Ba}]{zhou2022large}
Yongchao Zhou, Andrei~Ioan Muresanu, Ziwen Han, Keiran Paster, Silviu Pitis, Harris Chan, and Jimmy Ba. 2022.
\newblock Large language models are human-level prompt engineers.
\newblock \emph{arXiv preprint arXiv:2211.01910}.

\end{thebibliography}
\bibliographystyle{acl_natbib}

\newpage

\appendix
\section{Attack Experiment Details}
\label{sec:more_setting_details}
\subsection{Datasets}
\label{sec:appendix_dataset}
We provide the details for each  English generation dataset in this section. The base prompt template for each dataset is given in Table \ref{tab:prompt_format_dataset}.

For long-form question answering, we utilize ELI5~\cite{eli5_lfqa}. We choose the \texttt{reddit-eli5} split of HC3 as the training dataset to optimize attack prompts. The split includes human and ChatGPT answers for open-ended questions selected from ELI5. From the split, We collect each question as input instance, and the first human answer and a ChatGPT answer as output texts for train data. We remove the phrase \texttt{"Explain like I'm five"} in each question that does not exist in the original ELI5. We use the original ELI5 dataset as test set, after filtering out questions that also appear in the train set. 

For text generation, we use full news articles in XSum~\cite{narayan-etal-2018-dont} and Wikipedia articles in SQuAD~\cite{rajpurkar2016squad} as human writings. In each dataset, we ask the model to generate continuations of N=30 initial tokens in an article. For XSum, we follow the original train and test split. For SQuAD, we use the first half of the train set for optimizing attacks. We construct the test set by concatenating the validation set and the second half of the train set. We filter out the noisy non-English articles in the datasets with \texttt{en\_core\_web\_sm} model from spaCy~\cite{montani2023spacy} library.

\begin{table}
\centering
\begin{adjustbox}{width=1\linewidth}
\begin{tabular}{cl} 
\toprule
Dataset & \multicolumn{1}{c}{Prompt Template} \\ 
\midrule
ELI5 & \begin{tabular}[c]{@{}l@{}}Answer with at least 300 words.\\\\Question:\\\{question\}\\\\Answer:\end{tabular} \\ 
\midrule
\begin{tabular}[c]{@{}c@{}}SQuAD\\\&\\XSum\end{tabular} & \begin{tabular}[c]{@{}l@{}}Initial words:\\\{Initial 30 tokens\}\\\\Complete the article with at least \\
300 words, based on the initial words.\end{tabular} \\
\bottomrule                                                                                                                                                                                                
\end{tabular}
\end{adjustbox}
\caption{Base task description for each dataset.}
\label{tab:prompt_format_dataset}
\end{table}

\FloatBarrier

\begin{table*}
\centering
\small
\setlength{\tabcolsep}{6pt} 
\renewcommand{\arraystretch}{1} 
\begin{tabular}{cclccccccccc} 
\toprule
\multirow{2}{*}{Model} & \multirow{2}{*}{Metric} & \multirow{2}{*}{Attack} & \multicolumn{3}{c}{ChatGPT Detector} & \multicolumn{3}{c}{Perplexity} & \multicolumn{3}{c}{DetectGPT} \\
 &  &  & ELI5 & XSum & SQuAD & ELI5 & XSum & SQuAD & ELI5 & XSum & SQuAD \\ 
\midrule
\multirow{10}{*}{0301} & \multirow{5}{*}{\begin{tabular}[c]{@{}c@{}}N/A \\AUROC\end{tabular}} & PARA & 92.20 & 80.39 & 96.19 & 97.92 & 92.75 & 98.64 & 90.83 & 79.29 & 93.96 \\
 &  & DIPPER & 94.74 & 82.38 & 96.45 & 97.63 & 93.41 & 98.78 & 91.00 & 80.76 & 94.34 \\
 &  & SICO & 93.03 & 79.99 & 95.92 & 97.92 & 92.88 & 98.81 & 91.96 & 80.18 & 94.49 \\
 &  & IP & 93.29 & 80.49 & 95.91 & 97.98 & 93.03 & 98.81 & 91.63 & 80.21 & 94.43 \\
 &  & \attackMethod & 93.37 & 80.24 & 95.92 & 97.97 & 92.91 & 98.81 & 91.55 & 80.47 & 94.49 \\
\cmidrule{2-12}
 & \multirow{5}{*}{\begin{tabular}[c]{@{}c@{}}Best F1\end{tabular}} & PARA & 86.49 & 78.13 & 90.94 & 93.90 & 87.14 & 95.59 & 84.27 & 73.98 & 86.61 \\
 &  & DIPPER & 89.00 & 79.60 & 92.20 & 94.03 & 87.76 & 95.43 & 84.64 & 75.90 & 87.64 \\
 &  & SICO & 87.25 & 77.22 & 90.29 & 93.88 & 87.55 & 96.55 & 85.49 & 75.04 & 87.74 \\
 &  & IP & 87.46 & 77.62 & 90.29 & 94.03 & 87.74 & 96.55 & 85.67 & 75.23 & 86.89 \\
 &  & \attackMethod & 87.53 & 77.36 & 90.29 & 94.03 & 87.59 & 96.55 & 85.55 & 75.15 & 87.74 \\ 
\midrule
\multirow{10}{*}{0613} & \multirow{5}{*}{\begin{tabular}[c]{@{}c@{}}N/A\\AUROC\end{tabular}} & PARA & 98.12 & 86.05 & 91.73 & 97.11 & 87.24 & 96.98 & 91.77 & 79.54 & 92.62 \\
 &  & DIPPER & 98.78 & 86.93 & 92.27 & 97.47 & 87.18 & 95.93 & 92.48 & 79.95 & 92.07 \\
 &  & SICO & 98.02 & 86.10 & 91.73 & 97.14 & 87.19 & 96.98 & 91.22 & 79.27 & 92.62 \\
 &  & IP & 98.12 & 85.66 & 91.73 & 97.11 & 86.86 & 96.98 & 91.77 & 78.94 & 92.62 \\
 &  & \attackMethod & 98.12 & 86.05 & 91.73 & 97.11 & 87.24 & 96.98 & 91.77 & 79.54 & 92.62 \\
\cmidrule{2-12}
 & \multirow{5}{*}{\begin{tabular}[c]{@{}c@{}}Best F1\end{tabular}} & PARA & 86.49 & 78.13 & 90.94 & 93.90 & 87.14 & 95.59 & 84.92 & 75.68 & 85.56 \\
 &  & DIPPER & 89.00 & 79.60 & 92.20 & 94.03 & 87.76 & 95.43 & 85.96 & 75.72 & 84.64 \\
 &  & SICO & 87.25 & 77.22 & 90.29 & 93.88 & 87.55 & 96.55 & 84.90 & 75.66 & 85.56 \\
 &  & IP & 87.46 & 77.62 & 90.29 & 94.03 & 87.74 & 96.55 & 84.92 & 74.94 & 85.56 \\
 &  & \attackMethod & 87.53 & 77.36 & 90.29 & 94.03 & 87.59 & 96.55 & 84.92 & 75.68 & 85.56 \\
\bottomrule
\end{tabular}
\caption{Detection performance of generations for non-attack generations from gpt-3.5-turbo-0301 (0301) and gpt-3.5-turbo-0613 (0613).}
\label{tab:original_scores}
\end{table*}
\FloatBarrier
\subsection{Attack Implementations}
\label{sec:appendix_attack_methods}
We provide details about our implementations for the baseline attacks. See Table~\ref{tab:baseline_attack_prompts} for the revised prompts of baseline attacks in our implementations.

\begin{itemize}[noitemsep,topsep=1pt]
    \item \textbf{DIPPER} \cite{krishna2023paraphrasing} offers control codes to modify the extent of lexical changes (L) and reordering of contents (O). We use the harshest condition (L=60, O=60), and follow generation configurations of \citet{krishna2023paraphrasing} to paraphrase. 
    \item \textbf{PARA} We use random sampling with the temperature set as 1 to generate both the original and the paraphrased generations.
    \item \textbf{SICO} \cite{lu2024large} We follow the prompt templates in the official implementation of \citet{lu2024large}, with a small modification. The original template of SICO does not have a constraint on the length of outputs, leading to the generation of outputs shorter than the minimum length. To fix the issue, we insert a short phrase (\texttt{"using at least 300 words, "}) right after the common initial phrase (\texttt{"Based on the description, "}) of each task instruction, and append it at the end of the paraphrase instruction (\texttt{"Based on the description, rewrite this to P2 style answer"}) in the original prompts of \citet{lu2024large}. As we lengthen the outputs, the number of viable in-context examples decreases. We reduce the number of examples from 8 to 4, equal to the size of $B_{tr}$ in \attackMethod.
    
    \item \textbf{IP} \cite{shi2023red} 
    We update the base task descriptions in the original paper to fit our setting.
    
\end{itemize}

\begin{table*}[t]
\centering
\small
\setlength{\tabcolsep}{6pt} 
\renewcommand{\arraystretch}{1} 
\begin{tabular}{ccl} 
\toprule
Attack & Task & Attack Prompt \\ 
\midrule
PARA & - & \begin{tabular}[c]{@{}l@{}}Paraphrase this using at least 300 words.\\\\\{N/A generation\}\\\\Paraphrase:\end{tabular} \\ 
\midrule
\multirow{3}{*}{\begin{tabular}[c]{@{}c@{}}\\~\\~\\SICO\end{tabular}} & ELI5 & Based on the description, \textcolor{red}{using at least 300 words, }answer questions in P2 style writings \\ 
\cmidrule{2-3}
 & \begin{tabular}[c]{@{}c@{}}XSum\\\&\\SQuAD\end{tabular} & Based on the description, \textcolor{red}{using at least 300 words, }complete \textcolor{red}{the article }in P2 style writings: \\ 
\cmidrule{2-3}
 & \multicolumn{1}{l}{Paraphrase} & \begin{tabular}[c]{@{}l@{}}\{difference feature between human and AI\}\\Based on the description, rewrite this to P2 style writing \textcolor{red}{using at least 300 words}:\end{tabular} \\ 
\midrule
\multirow{2}{*}{\begin{tabular}[c]{@{}c@{}}\\~\\~\\IP\end{tabular}} & ELI5 & \begin{tabular}[c]{@{}l@{}}\textcolor{red}{Answer }with at least \textcolor{red}{300 }words.\\Question:\\\{question\}\\\\\{prompt\}\\Answer:\end{tabular} \\ 
\cmidrule{2-3}
 & \begin{tabular}[c]{@{}c@{}}XSum\\\&\\SQuAD\end{tabular} & \begin{tabular}[c]{@{}l@{}}\textcolor{red}{Initial words:}\\\{question\}\\\\\textcolor{red}{Complete the article with at least 300 words, based on the initial words.}\\\{prompt\}\end{tabular} \\
\bottomrule
\end{tabular}
\caption{Prompts for each attack utilized in our experiments. The prompt for PARA with the "-" task is applied to all tasks. The "Paraphrase" prompt in SICO refers to the prompt used in every task to initialize the in-context examples. We represent our modifications from the original paper in red.}
\label{tab:baseline_attack_prompts}
\end{table*}

\subsection{Performances on Non-Attack Generations}
Different attack experiments share the non-attack (N/A) generations if they share the generation model and task. However, as the lengths of attack generations vary, human texts and N/A generations are truncated on different locations. Therefore, we ensure the validity of the experiment, we report the detection scores for non-attack generations on each test in Table~\ref{tab:original_scores}. We observe a small variance in AUROC due to the truncation, but it is negligible to the drops attack generations cause in Table \ref{tab:attack_0301} and \ref{tab:attack_0613}.

\begin{table*}
\centering
\begin{tabular}{cl} 
\toprule
Name & \multicolumn{1}{c}{Prompt} \\ 
\midrule
$p_{disc}$ & \begin{tabular}[c]{@{}l@{}}G1's writing \#1.\\\{Human text\}\\...\\\\G1's writing \#\{Number of text pairs\}\\\{Human text \#\{Number of text pairs\}\}\\\\G2's writing \#1.\\\{AI text \#1\}\\...\\\\G2's writing \#\{Number of text pairs\}.\\\{AI text \#\{Number of text pairs\}\}\\\\Provide a list containing \{feedback\_list\_length\} general, representative characteristics \\of G1's writings compared to G2's writings.\\\\List of \{feedback\_list\_length\} characteristics:\end{tabular} \\ 
\midrule
$p_{ins}$ & \begin{tabular}[c]{@{}l@{}}You are a helpful assistant that generate brief instructions to help others write like \\G1's answers. You will be provided with a list of feedbacks. Convert each feedback \\to a brief instruction asking you to write like G1's answers. Only mention what to do in \\each instruction. Do not mention 'G1' or 'G2' in the instructions.\\Feedbacks:\\\{feedback\}\end{tabular} \\ 
\midrule
$p_{MC}$ & \begin{tabular}[c]{@{}l@{}}Generate a variation of the input instruction while keeping the semantic meaning.\\\\Input:\\\{mc\_feedback\}\\\\Output:\end{tabular} \\
\bottomrule               
\end{tabular}
\caption{Pre-defined prompts for the optimization process of \attackMethod}
\label{tab:prompt_format_attack}
\end{table*}

\FloatBarrier

\begin{table*}
\centering
\begin{tabular}{cl} 
\toprule
Task & \multicolumn{1}{c}{Prompt Template} \\ 
\midrule
Revision & \begin{tabular}[c]{@{}l@{}}You will be given a question and a major difference between human and ChatGPT.\\Your task is to write a human-like answer.\\Please make sure you read and understand these instructions carefully.\\Major Difference between human and ChatGPT:\\\{A human annotation of major difference\}\\Q: \{Question\}\\A:\end{tabular} \\ 
\midrule
Judge & \begin{tabular}[c]{@{}l@{}}You will be given two answers written for the same question.\\Your task is to find the most human-like answer.\\Please make sure you read and understand these instructions carefully.\\Evaluation Criteria:\\\{A human annotation of major difference\}\\Answer 1:\\\{Answer 1\}\\Answer 2:\\\{Answer 2\}\\Human-like answer:\end{tabular} \\
\bottomrule
\end{tabular}
\caption{Prompt templates for assessing the existence of prompt-specific shortcut features on HC3.}
\label{tab:prompt_format_existence}
\end{table*}
\FloatBarrier

\begin{algorithm*}[t]
\caption{\attackMethodFull}
\label{alg:attack_algorithm}
\small
\textbf{Input}: Train data \textbf{$D_{tr}$}, Validation data \textbf{$D_{val}$}, initial prompt \textbf{$p_{0}$}\\
\textbf{Parameter}: Generative model $G$, Beam size $k$, maximum train step $step_{max}$, pre-defined manual prompts $p_{disc}, p_{ins}, p_{MC}$\\
\textbf{Output}: Optimal instruction list $i_{opt}$
\begin{algorithmic}[1] 
\STATE $I \gets \{i_{0}\}$
\FOR {$step = 1, \cdots , step_{max}$}
    \STATE Sample minibatch $B_{tr}=\{x_{tr}^{m}, h_{tr}^{m}\}_{m=1}^{N_{tr}}$, $B_{val}=\{x_{val}^{n}, h_{val}^{n}\}_{n=1}^{N_{val}}$ from $D_{tr}$ and $D_{val}$
    \STATE $I_{inter} \gets \emptyset$
    \FORALL {$i_{curr} \in I$}
        \STATE Generate AIGC from current instructions $Y_{curr}= \{y_{curr}^{m}\}_{m=1}^{N_{tr}}$, where $y_{curr}^{m}=G_(t,i_{curr},x_{tr}^{m}) $
    
        \STATE Get a feedback list of $N_{feed}$ items $L_{feed} \gets G(p_{disc}, h_{tr}^{1}\oplus ... \oplus h_{tr}^{N_{tr}}, y_{curr}^{1}\oplus \cdots \oplus y_{curr}^{N_{tr}})$
        
        \STATE Construct candidate instructions from each feedback item $I_{cand} \gets G(t, p_{ins}, l_{feed}^{m}), \forall{m} \in \{1,\cdots,N_{feed}\} $
        \STATE $I_{inter} \gets I_{inter} \oplus getTopK(B_{val}, I_{cand}) $
    \ENDFOR
    \STATE Get paraphrased candidates $I_{MC} \gets G(p_{MC}, \emptyset, i_{inter}^{k}) \in I$
    \STATE $I \gets getTopK(B_{val}, I_{MC}  \oplus I_{inter}) $
\ENDFOR
\STATE \textbf{return} Optimized adversarial instruction list $i_{opt} \gets I[0]$
\end{algorithmic}
\end{algorithm*}

\FloatBarrier

\begin{algorithm}[t]
\caption{getTopK}
\label{alg:evaluation_and_select}
\small
\textbf{Input}: Evaluation data batch \textbf{$B_{val}=\{x_{val}^{n}, h_{val}^{n}\}_{n=1}^{N_{val}}$}, a set of candidate instructions $I_{cand}=\{i_{cand}\}_{n=1}^{N_{cand}}$ \\
\textbf{Parameter}: Generative model $G$, basic task description $t$\\
\textbf{Output}: top-k adversarial instructions sorted in the descending order $I_{best}$
\begin{algorithmic}[1] 

\STATE Collect generations from each input instance

$G_{i,j} \gets G(t, i_{cand}^{i}, x_{val}^{j}), $ where

$\forall{i} \in \{1, \cdots, N_{cand} \}, \forall{j} \in \{1, \cdots, N_{val} \}$


\STATE sort $I_{cand}$ in the descending order of $score(G_{i})$,  where $ score(Y) = \frac{1}{N} \displaystyle\sum_{i=0} ^{|Y|} \mathbb{1} (f(Y_{i}) \geq \tau ) $ 

\STATE \textbf{return} top-k adversarial instruction $I_{best} \subseteq I_{cand}$

\end{algorithmic}
\end{algorithm}

\section{\attackMethod\ Implementation}
\label{sec:appendix_algo}
We illustrate the pseudo code of \attackMethod\ in Algorithm~\ref{alg:attack_algorithm} and \ref{alg:evaluation_and_select}. 
In the algorithm, there are several lines where we provide pre-defined manual prompts to the LLM. We provide the manual prompts in Table \ref{tab:prompt_format_attack}.

\section{Full Robustness Evaluation Results}
\label{sec:augmented_full}
Table \ref{tab:robust_0301_human_score} and \ref{tab:robust_0613_human_score} present the human scores of various attack generations from gpt-3.5-turbo-0301 (0301) and gpt-3.5-turbo-0613 (0613), respectively. \textbf{Full} generally achieves the lowest human score in every setting, and we find that the monotonic decrease of the human score on gpt-3.5-turbo-0613 generations also only appears in gpt-3.5-turbo-0301, except for \attackMethod. The increase on \attackMethod\ is still small and \textbf{Full} shows better scores than other detectors on \attackMethod.

\begin{table}[t!]
\centering
\small
\setlength{\tabcolsep}{3pt} 
\renewcommand{\arraystretch}{1} 
\begin{adjustbox}{width=1\linewidth}
\begin{tabular}{l|ccccc||cc} 
\toprule
     & Fi.  & Med. & QA   & ELI5 & CSAI & \begin{tabular}[c]{@{}c@{}}Macro\\Avg.\end{tabular} & \begin{tabular}[c]{@{}c@{}}Micro\\Avg.\end{tabular}  \\ 
\midrule
div.   & 0.71 & 0.75 & 0.85 & 0.88 & 0.65 & 0.77       & 0.77        \\
subj.  & 0.29 & 0.37 & 0.49 & 0.73 & 0.33 & 0.44       & 0.47        \\
cas. & 0.94 & 0.86 & 0.94 & 0.97 & 0.99 & 0.94       & 0.95        \\
emo.   & 0.31 & 0.69 & 0.31 & 0.90 & 0.08 & 0.46       & 0.46        \\
\bottomrule
\end{tabular}
\end{adjustbox}
\caption{The preference of GPT-4 to the answers guided with human annotations.}
\label{tab:existence_experiment}
\end{table}
\FloatBarrier

\begin{table*}[t]
\centering
\small
\setlength{\tabcolsep}{6pt} 
\renewcommand{\arraystretch}{1} 
\begin{tabular}{cl|c|ccc|ccc|ccc} 
\toprule
\multicolumn{2}{c|}{\multirow{2}{*}{}} & No train & \multicolumn{3}{c|}{Full} & \multicolumn{3}{c|}{- N/A} & \multicolumn{3}{c}{- \attackMethod} \\
\multicolumn{2}{c|}{} &  & 500 & 1000 & 2000 & 500 & 1000 & 2000 & 500 & 1000 & 2000 \\ 
\midrule
\multirow{3}{*}{N/A} & ELI5 & 10.77 & 1.82 & 1.03 & 0.70 & 3.42 & 10.78 & 11.39 & 3.02 & 1.11 & 1.25 \\
 & XSum & 14.38 & 3.00 & 1.45 & 1.02 & 7.45 & 13.67 & 13.63 & 5.54 & 2.08 & 2.78 \\
 & SQuAD & 3.34 & 1.72 & 0.72 & 0.42 & 2.88 & 10.38 & 10.65 & 2.25 & 0.68 & 0.92 \\ 
\midrule
\multirow{3}{*}{PARA} & ELI5 & 18.11 & 1.88 & 0.75 & 0.45 & 4.16 & 10.83 & 11.41 & 1.96 & 0.76 & 0.83 \\
 & XSum & 33.26 & 2.98 & 0.92 & 0.64 & 7.39 & 13.94 & 12.70 & 2.79 & 1.03 & 1.15 \\
 & SQuAD & 19.11 & 1.93 & 0.78 & 0.44 & 3.59 & 10.83 & 11.43 & 1.73 & 0.71 & 0.89 \\ 
\midrule
\multirow{3}{*}{DIPPER} & ELI5 & 17.17 & 6.20 & 9.33 & 10.86 & 26.30 & 30.57 & 35.82 & 12.60 & 7.64 & 12.13 \\
 & XSum & 13.30 & 6.33 & 8.47 & 7.13 & 19.78 & 26.10 & 28.50 & 16.42 & 9.42 & 14.70 \\
 & SQuAD & 20.86 & 8.21 & 11.30 & 13.25 & 28.08 & 34.81 & 40.23 & 13.64 & 7.31 & 16.91 \\ 
\midrule
\multirow{3}{*}{SICO} & ELI5 & 35.08 & 6.28 & 7.21 & 3.86 & 5.02 & 15.48 & 13.52 & 64.95 & 46.27 & 48.21 \\
 & XSum & 48.73 & 3.78 & 2.56 & 2.07 & 4.66 & 13.77 & 15.67 & 30.94 & 26.84 & 28.33 \\
 & SQuAD & 45.37 & 1.66 & 0.70 & 0.42 & 2.43 & 9.88 & 10.99 & 1.71 & 0.73 & 0.87 \\ 
\midrule
\multirow{3}{*}{IP} & ELI5 & 18.11 & 1.81 & 0.72 & 0.44 & 2.58 & 10.25 & 10.77 & 2.24 & 0.79 & 0.97 \\
 & XSum & 56.33 & 12.80 & 9.76 & 6.94 & 36.49 & 38.40 & 33.06 & 17.87 & 8.09 & 11.52 \\
 & SQuAD & 29.13 & 2.81 & 0.94 & 0.65 & 3.96 & 11.95 & 11.85 & 2.86 & 1.00 & 1.36 \\ 
\midrule
\multirow{3}{*}{\attackMethod} & ELI5 & 43.63 & 1.66 & 0.93 & 0.50 & 1.27 & 9.12 & 5.06 & 56.53 & 39.43 & 42.79 \\
 & XSum & 28.81 & 5.03 & 3.77 & 5.14 & 8.44 & 15.12 & 15.31 & 36.30 & 33.58 & 34.45 \\
 & SQuAD & 12.63 & 1.87 & 0.80 & 0.47 & 3.09 & 10.41 & 9.70 & 2.03 & 0.58 & 1.02 \\
\bottomrule
\end{tabular}
\caption{Human score of the original and additionally trained detectors on ChatGPT (gpt-3.5-turbo-0301) generations. We present each score in percentage.}
\label{tab:robust_0301_human_score}
\end{table*}

\begin{table*}[t]
\centering
\small
\setlength{\tabcolsep}{6pt} 
\renewcommand{\arraystretch}{1} 
\begin{tabular}{cl|c|ccc|ccc|ccc} 
\toprule
\multicolumn{2}{c|}{\multirow{2}{*}{}} & No train & \multicolumn{3}{c|}{Full} & \multicolumn{3}{c|}{- N/A} & \multicolumn{3}{c}{- \attackMethod} \\
\multicolumn{2}{c|}{} &  & 500 & 1000 & 2000 & 500 & 1000 & 2000 & 500 & 1000 & 2000 \\ 
\midrule
\multirow{3}{*}{N/A} & ELI5 & 2.19 & 1.41 & 0.70 & 0.43 & \multicolumn{1}{c}{1.97} & \multicolumn{1}{c}{9.31} & \multicolumn{1}{c|}{9.73} & \multicolumn{1}{c}{1.98} & \multicolumn{1}{c}{0.80} & \multicolumn{1}{c}{0.87} \\
 & XSum & 6.28 & 1.45 & 0.67 & 0.43 & \multicolumn{1}{c}{1.06} & \multicolumn{1}{c}{9.01} & \multicolumn{1}{c|}{9.12} & \multicolumn{1}{c}{2.34} & \multicolumn{1}{c}{0.72} & \multicolumn{1}{c}{0.86} \\
 & SQuAD & 6.02 & 1.40 & 0.67 & 0.42 & \multicolumn{1}{c}{1.43} & \multicolumn{1}{c}{9.21} & \multicolumn{1}{c|}{9.47} & \multicolumn{1}{c}{1.68} & \multicolumn{1}{c}{0.71} & \multicolumn{1}{c}{0.86} \\ 
\midrule
\multirow{3}{*}{PARA} & ELI5 & 4.53 & 1.32 & 0.70 & 0.43 & 1.38 & 9.05 & 9.28 & 1.81 & 0.70 & 0.82 \\
 & XSum & 6.49 & 1.36 & 0.68 & 0.43 & 0.88 & 8.94 & 9.08 & 1.87 & 0.71 & 0.83 \\
 & SQuAD & 8.11 & 1.35 & 0.69 & 0.43 & 1.12 & 9.06 & 9.23 & 1.63 & 0.69 & 0.81 \\ 
\midrule
\multirow{3}{*}{DIPPER} & ELI5 & 15.64 & 4.96 & 5.90 & 6.64 & 21.68 & 26.19 & 32.51 & 8.45 & 4.82 & 8.12 \\
 & XSum & 12.51 & 3.59 & 3.32 & 3.38 & 9.30 & 16.75 & 17.84 & 9.48 & 4.60 & 8.32 \\
 & SQuAD & 25.22 & 6.12 & 7.97 & 9.07 & 19.72 & 27.43 & 31.44 & 11.35 & 6.53 & 13.60 \\ 
\midrule
\multirow{3}{*}{SICO} & ELI5 & 24.71 & 2.14 & 1.13 & 1.22 & 1.20 & 9.23 & 9.25 & 92.14 & 88.11 & 89.90 \\
 & XSum & 12.89 & 1.35 & 0.71 & 0.45 & 0.91 & 9.10 & 9.17 & 1.75 & 0.71 & 0.83 \\
 & SQuAD & 73.05 & 1.45 & 0.70 & 0.43 & 1.45 & 8.96 & 9.18 & 2.45 & 0.87 & 1.21 \\ 
\midrule
\multirow{3}{*}{IP} & ELI5 & 10.03 & 1.41 & 0.69 & 0.43 & 1.26 & 9.11 & 9.39 & 2.02 & 0.75 & 0.85 \\
 & XSum & 17.66 & 1.41 & 0.69 & 0.43 & 0.90 & 8.95 & 9.08 & 2.30 & 0.93 & 1.06 \\
 & SQuAD & 10.57 & 1.60 & 0.89 & 0.62 & 2.04 & 9.69 & 9.90 & 2.35 & 0.93 & 1.22 \\ 
\midrule
\multirow{3}{*}{\attackMethod} & ELI5 & 79.23 & 1.38 & 0.75 & 0.42 & 0.96 & 8.90 & 9.06 & 24.46 & 10.77 & 14.62 \\
 & XSum & 22.57 & 1.73 & 0.75 & 0.45 & 1.53 & 9.26 & 9.22 & 12.45 & 4.97 & 7.43 \\
 & SQuAD & 71.23 & 1.47 & 0.69 & 0.46 & 0.92 & 8.92 & 9.11 & 7.19 & 2.24 & 4.14 \\
\bottomrule
\end{tabular}
\caption{Human score of the original and additionally trained detectors on ChatGPT (gpt-3.5-turbo-0613) generations. We present each score in percentage.}
\label{tab:robust_0613_human_score}
\end{table*}
\section{Analyzing Prompt-Specific Features in Train Data}
\label{sec:HC3_analysis}

In this section, we empirically find prompt-specific features in HC3 and show their relevance to \attackMethod\ instructions. 

\subsection{Existence of Shortcuts}
\label{sec:HC3_shorcut_existence}

\subsubsection{Setting}
\label{sec:existence_setting}
\paragraph{Subject Dataset}
We test the existence of prompt-specific shortcuts in Human ChatGPT Comparison Corpus (HC3) \citep{guo-etal-2023-hc3}. HC3 consists of ChatGPT and human answers from five different tasks, namely finance (\textbf{Fi.}), medicine (\textbf{Med.}), open\_qa (\textbf{QA}), reddit\_eli5 (\textbf{ELI5}), and wiki\_csai (\textbf{CSAI}). \citet{guo-etal-2023-hc3} provides a summary of four major differences between the writings of two author groups in the dataset. We name the difference annotations in order of appearance in \citet{guo-etal-2023-hc3}: 1.~diversity (\textbf{div.}), 2.~subjectivity (\textbf{subj.}), 3.~casualness (\textbf{cas.}), and 4.~emotionality (\textbf{emo.}). We utilize them as our difference annotations without any modifications and check if there are prompt-specific features among them. Refer to \citet{guo-etal-2023-hc3} for full annotations. 

\paragraph{Finding Shortcuts}
For each task in HC3, we select 100 questions and generate answer with ChatGPT (gpt-3.5-turbo-0301). From the 500 questions, we filter out the questions that ChatGPT refused to answer, and 394 questions remain. Then, for each remaining question, we ask the model to generate revised answers providing one of four human annotations of the major difference between human and ChatGPT.

We compare the generations from the different prompts to verify if ChatGPT can adjust its behavior with distinctive human characteristics. To this end, we utilize GPT-4~\cite{achiam2023gpt} as a judge to evaluate which answer fits the description of a human feature better. Specifically, GPT-4 receives two ChatGPT answers, where each answer is generated with or without the description of the difference, and we ask GPT-4 to pick an answer closer to human, concerning the description of the difference that ChatGPT used. The order of two answers is randomized to remove the effect of inherent order bias in GPT-4. Our prompt template for this experiment is given in Table \ref{tab:prompt_format_existence}.

\subsubsection{Experiment Result \& Discussion}

Table~\ref{tab:existence_experiment} shows the proportion of the cases where GPT-4 favored ChatGPT answers guided with the additional prompt. We find that with corresponding prompts, ChatGPT could tweak the outputs to better align the answers with human features. For diversity and casualness, the revised answers are preferred in every task. The revision on casualness achieves the win ratio of 0.94, proving that the impact of instructions can be severe. Overall, this shows that the previous human analysis, and the dataset itself, do not represent the prompt-invariant features of the model.

\subsection{Comparison to \attackMethod\ Instructions}
We observe the efficacy of \attackMethod\ in finding deceptive instructions that perturb prompt-specific features. To confirm that such weakness resulted from the bias in train data, we collect the 82 instructions from final \attackMethod\ instruction lists from the 18 \attackMethod\ runs in Section \ref{sec:attack}, and compare their contents to the human annotations of major differences in \citet{guo-etal-2023-hc3}.

We consistently find instructions related to the features of the train data from each run, proving that \attackMethod\ successfully exploits prompt-specific features and ChatGPT detector depends on decision rules related to the data collection prompts of HC3. We present the example \attackMethod\ instructions relevant to one of the major difference annotations from \citet{guo-etal-2023-hc3} ijn Table \ref{tab:instruction_annonation_comparison}.

\begin{table*}[t]
\centering
\small
\setlength{\tabcolsep}{6pt} 
\renewcommand{\arraystretch}{1} 
\begin{adjustbox}{width=1\textwidth}
\begin{tabular}{cl} 
\toprule
AI Feature & \multicolumn{1}{c}{Relevant  \attackMethod Instructions} \\ 
\midrule
diversity & \begin{tabular}[c]{@{}l@{}}-~­Direct your responses towards particular occurrences or undertakings.\\-~­Provide more background information and context in your answers.\\-~­Offer responses that commonly refer to historical occurrences or background information.\end{tabular} \\ 
\midrule
subjectivity & \begin{tabular}[c]{@{}l@{}}-~­Incorporate exact quotations from news outlets into your responses.\\-~­Make sure to incorporate quotes or references from historical sources when formulating your responses.\\-~­Include quotes and references from experts in your answers.\\-~­Make sure to cite sources and authors in your responses.\\-~­When answering, try to include quotes from individuals who were present at the event or involved\\
\ \ in the story.\end{tabular} \\ 
\midrule
casualness & \begin{tabular}[c]{@{}l@{}}­-~Incorporate humorous or sarcastic elements to captivate the reader in your responses.\\-~­Incorporate witty remarks and irony to convey your message in your responses.\\-~­Please include humor or lightheartedness in your answers.\\-~­Respond using wit or irony.\\-~Respond using a more humorous or casual tone.\\-~­Use informal language and tone in your answers.\end{tabular} \\ 
\midrule
emotionality & \multicolumn{1}{c}{-} \\
\bottomrule
\end{tabular}
\end{adjustbox}
\caption{Major distinctive features of ChatGPT detector in HC3, and \attackMethod\ instructions correspondent with each feature. We do not provide instructions relevant to emotionality as we did not such instructions.}
\label{tab:instruction_annonation_comparison}
\end{table*}

\end{document}